\documentclass[11pt]{article}

\usepackage[final]{acl}

\usepackage{times}
\usepackage{latexsym}

\usepackage[T1]{fontenc}

\usepackage[utf8]{inputenc}

\usepackage{microtype}

\usepackage{inconsolata}

\usepackage{graphicx}

\usepackage{subfiles} 
\usepackage{multicol}
\usepackage{amsmath}
\usepackage{booktabs}
\usepackage{amsfonts}
\usepackage{amssymb}
\usepackage{booktabs}
\usepackage{multirow}
\usepackage{subcaption}
\usepackage{adjustbox}
\newcommand{\kv}{ZoomR}
\usepackage{algorithm}
\usepackage{algpseudocode}
\usepackage{subcaption}
\usepackage{tcolorbox}
\usepackage{bm}

\usepackage{color}

\title{ZoomR: Memory Efficient Reasoning through  Multi-Granularity Key Value Retrieval}

\author{
  David H. Yang\textsuperscript{1}, Yuxuan Zhu\textsuperscript{1},  Mohammad Mohammadi Amiri\textsuperscript{1}, \\ 
  \textbf{Keerthiram Murugesan\textsuperscript{2},  Tejaswini Pedapati\textsuperscript{2} 
  , Subhajit Chaudhury\textsuperscript{2}, Pin-Yu Chen\textsuperscript{2}} \\
  \textsuperscript{1}Rensselaer Polytechnic Institute \quad \textsuperscript{2}IBM Research\\
  {\tt \{yangd13, zhuy27, mamiri\}@rpi.edu} \quad {\tt tejaswinip@us.ibm.com} \\
  {\tt \{keerthiram.murugesan, subhajit, pin-yu.chen\}@ibm.com}
}

\begin{document}
\maketitle
\begin{abstract}
Large language models (LLMs) have shown great performance on complex reasoning tasks but often require generating long intermediate thoughts before reaching a final answer.\ During generation, LLMs rely on a key-value (KV) cache for autoregressive decoding.\ However, the memory footprint of the KV cache grows with output length.\ Prior work on KV cache optimization mostly focus on compressing the long input context, while retaining the full KV cache for decoding.\ For tasks requiring long output generation, this leads to increased computational and memory costs.\ In this paper, we introduce ZoomR, a novel approach that enables LLMs to adaptively compress verbose reasoning thoughts into summaries and uses a dynamic KV cache selection policy that leverages these summaries while also strategically ``zooming in'' on fine-grained details.\ By using summary keys as a coarse-grained index during decoding, ZoomR uses the query to retrieve details for only the most important thoughts.\ This hierarchical strategy significantly reduces memory usage by avoiding full-cache attention at each step.\ Experiments across math and reasoning tasks show that our approach achieves competitive performance compared to baselines, while reducing inference memory requirements by more than $4\times$.\ These results demonstrate that a multi-granularity KV selection enables more memory efficient decoding, especially for long output generation.
\end{abstract}

\section{Introduction}

Large language models (LLMs) have demonstrated remarkable proficiency in complex reasoning tasks such as mathematics and coding \citep{openai2024openaio1card,deepseekai2025deepseekr1incentivizingreasoningcapability}.\ This success can be attributed to their ability to generate long chain of thoughts before arriving at a final solution.\ However, this verbosity introduces a significant computational and memory bottleneck.\ The autoregressive decoding process relies on a key-value (KV) cache that stores hidden states from previously generated tokens to enable efficient self-attention, and this cache grows linearly with the sequence length.\ This presents two critical challenges for long-output generation.\ First, the KV cache consumes a large amount of GPU memory, and second, the computational cost of attending to the ever-growing cache makes each subsequent token progressively slower to generate.\ For instance, generating a 16K-token response with a Llama-3.1-8B \citep{grattafiori2024llama3herdmodels} model for a batch size of 8 requires over 16GB of GPU memory for the KV cache, making deployment more challenging on consumer GPUs with less memory.

To address these challenges, several efficient reasoning strategies have been proposed \citep{sui2025stopoverthinkingsurveyefficient}.\ One line of work focuses on enabling the LLM to generate fewer tokens using techniques like prompt engineering \citep{han2025tokenbudgetawarellmreasoning,aytes2025sketchofthoughtefficientllmreasoning}, specialized training algorithms \citep{kimiteam2025kimik15scalingreinforcement,xia2025tokenskipcontrollablechainofthoughtcompression}, and latent space reasoning \citep{hao2024traininglargelanguagemodels,shen2025efficientreasoninghiddenthinking}.\ For these works, the primary objective is to speed up inference, and may implicitly reduce GPU memory consumption.\ Another direction, which we follow in this paper, aims to optimize the KV cache itself.\ While several methods for cache eviction or compression exist, most are optimized for long input contexts during the prefill stage, when the model processes the prompt, \citep{li2024snapkv,tang2024questqueryawaresparsityefficient,sun2025shadowkvkvcacheshadows,zhu2025sentencekvefficientllminference}, rather than addressing the unqiue memory challenges of long-output generation during decoding.\ Existing dynamic KV cache methods lose critical long-range information in long-output generation, resulting in poor performance on complex reasoning tasks \citep{xiao2024efficientstreaminglanguagemodels,zhang2023h2oheavyhitteroracleefficient,chen2025sepllmacceleratelargelanguage}.\ This leaves a crucial gap for an efficient cache management strategy tailored specifically for lengthy, generated reasoning chains, while maintaining high performance.

To address this gap, we propose ZoomR, a method inspired by established principles of human memory and attention.\ When solving complex problems, humans leverage hierarchical memory representations, maintaining schematic overviews of past information while selectively retrieving detailed specifics only when contextually relevant \citep{bartlett1932remembering}.\ ZoomR applies this strategy to LLMs by maintaining both compressed summaries (coarse-grained representations) of the generation history and the original detailed text, accessing fine-grained information only when the current context demands higher fidelity.\ We first fine-tune reasoning models to generate summaries of thoughts after each paragraph.\ During inference, ZoomR uses the current query and summary keys across all attention heads to identify the most important segments of the past thoughts.\ Then, a consensus is performed to retrieve the most important full-resolution details.\ By doing so, ZoomR only selects a small number of KVs to keep on GPU for attention computation, while offloading the full KV cache to CPU.\ Our implementation involves a one-time fine-tuning step to teach the model to summarize its own history, followed by a training-free dynamic cache selection policy at inference time.

Our contributions are three-fold.\ First, we introduce ZoomR, a dynamic, multi-granularity KV cache management technique designed to enable memory efficient long-output reasoning.\ Second, we present a practical framework that combines lightweight fine-tuning for summary generation with an efficient, approximate attention score-based retrieval mechanism for inference.\ Third, we empirically validate ZoomR on reasoning models, Qwen and Llama \citep{qwen2025qwen25technicalreport, grattafiori2024llama3herdmodels}, across challenging reasoning benchmarks, AIME2025 and MATH500.\ Our results demonstrate that ZoomR achieves performance comparable to a standard full KV cache while saving more than $4\times$ the GPU memory usage during inference.

\section{Background}

We investigate the efficiency of autoregressive inference for LLMs in long-generation reasoning tasks.\ In this setting, the model is given a prompt $\bm{X}$ of length $N_p$ and is tasked with generating a lengthy response $\bm{Y}$ of length $N_g$, $N_g \gg N_p$.\ Modern reasoning models trained using chain-of-thought prompting significantly improve performance over standard instruction-tuned models.\ However, this improvement comes at the cost of generating much longer sequences.\ The output $\bm{Y}$ is often structured into a detailed reasoning or ``thought'' section, followed by a final ``solution'' section, where the length of the thought process is typically much greater than that of the final answer.\ With recent advancements, models are capable of generating outputs that extend to tens of thousands of tokens.

At the core of the Transformer architecture is the self-attention mechanism, which is central to the decoding process.\ During autoregressive generation, a token is produced at each timestep $t$.\ The model has $N_{L}$ layers and $H$ attention heads.\ Let $\bm{x}_t \in \mathbb{R}^{d_\text{emb}}$ be the input embedding for the token.\ For a given position $t$, layer $l$, and head $h$, the input $\bm{x}_t$ is projected into query, key, and value vectors  $\bm{q}_{t}$, $\bm{k}_{t}$, and $\bm{v}_{t}$, of dimension $d$:
\begin{equation*}
    \bm{q}_t = \bm{x}_t \bm{W}_q, \quad \bm{k}_t = \bm{x}_t \bm{W}_k, \quad \bm{v}_t = \bm{x}_t \bm{W}_v,
\end{equation*}
where $\bm{W}_q, \bm{W}_k, \bm{W}_v \in \mathbb{R}^{d_\text{emb} \times d}$ are learned projection matrices.\ For ease of notation, we have dropped the index for layers and heads.\ To maintain context from all previous tokens, both from the prompt and the ongoing generation, the keys and values from all past steps are stored in a KV cache.\ At each step $t$, the newly computed key $\bm{k}_t$ and value $\bm{v}_t$ are appended to the cached matrices $\bm{K}_{t-1}$ and $\bm{V}_{t-1}$:
\begin{equation*}
    \bm{K}_t = [\bm{K}_{t-1} ; \bm{k}_t], \quad \bm{V}_t = [\bm{V}_{t-1} ; \bm{v}_t],
\end{equation*}
where $\bm{K}_t, \bm{V}_t  \in \mathbb{R}^{(N_p + t)\times d}$.
The attention output $\bm{A}$ is then computed as the scaled dot product between the query and cached keys, followed, by a softmax operation to obtain attention weights, and using them to aggregate the corresponding values:
\begin{equation*}
    \bm{A}(\bm{q}_t, \bm{K}_t, \bm{V}_t) 
    = \text{softmax}\!\left(\frac{\bm{q}_t \bm{K}_t^\top}{\sqrt{d}}\right) \bm{V}_t \in \mathbb{R}^{d}. 
\end{equation*}

This standard decoding process presents two major efficiency challenges, particularly as the generation length grows.\ First, the computational cost of generating each new token scales with the sequence length.\ The matrix-vector multiplication $\bm{q}_t \bm{K}_t^\top$ scales linearly with the length of the KV cache, resulting in quadratic complexity with respect to the total generation length.\ As the generation proceeds and the cache grows, each subsequent token becomes progressively more expensive to compute, making the overall inference process slow for long outputs.\ The second challenge is in GPU memory requirements.\ The primary memory bottleneck during LLM inference is the KV cache.\ The size of the cache matrices, $\bm{K}_t$ and $\bm{V}_t$, grows linearly with the sequence length.\ For a model with multiple layers and attention heads, this memory footprint can become prohibitively large, limiting the maximum sequence length that can be processed on available hardware.

\section{Methodology}
In this section, we present ZoomR, a method that enables memory-efficient reasoning through multi-granularity KV retrieval.\ We begin by motivating our approach followed by its two main components: (1) fine-tuning an LLM to learn how to generate summaries on the fly, and (2) a dynamic KV cache selection policy for memory efficient decoding.

\subsection{Motivation}
As LLM generation length increases, the KV cache creates practical limitations for extended reasoning tasks.\ Existing approaches attempt to address this challenge through various strategies, but often introduce significant limitations that compromise reasoning quality.\ Streaming-based approaches like StreamingLLM \citep{xiao2024efficientstreaminglanguagemodels}, retain attention sinks and a sliding window of recent KV cache entries, but suffer from the ``lost-in-the-middle'' problem, where critical intermediate reasoning steps are permenantly discarded.\ Dynamic token selection methods like H2O \citep{zhang2023h2oheavyhitteroracleefficient} preserve important tokens through attention-based scoring mechanisms.\ However, these approaches face two critical limitations.\ First, computing token-level importance scores at each generation step incurs substantial computational overhead.\ Second, token-level selection operates at a granularity that may miss semantic coherence, potentially fragmenting coherent reasoning segments.

\begin{figure}[!t]
\centering
\includegraphics[width=\columnwidth]{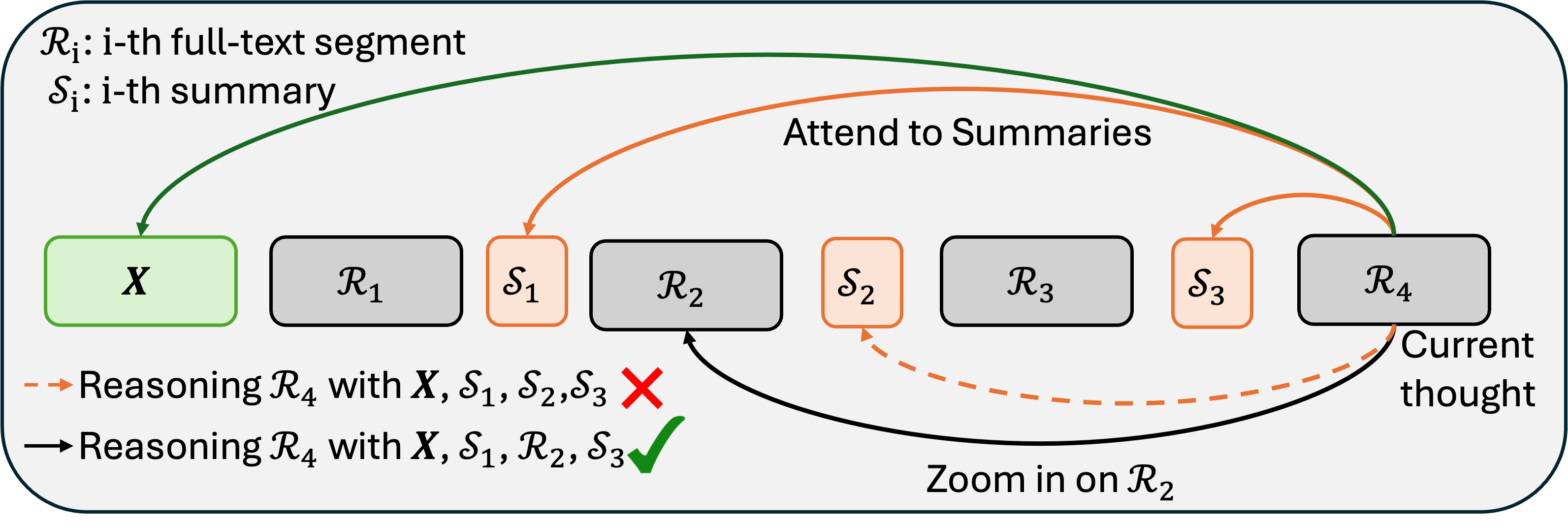}
\caption{Fine-grained details are often needed in order to generate the correct reasoning.\ Attending to only a compressed source of information like summaries can lead to incorrect intermediate steps.}
\label{fig2}
\end{figure}

Recent work has also explored using compressed summaries to reduce context size while preserving semantic information \citep{zhang2025lightthinkerthinkingstepbystepcompression,yan2025inftythinkbreakinglengthlimits}.\ While summarization provides meaningful compression, restricting attention exclusively to summary representations creates an information bottleneck.\ The key insight driving our approach is that different reasoning steps require access to historical context at different levels of granularity (see Figure \ref{fig2}).\ When tackling a challenging mathematical problem, the solution can draw on multiple lemmas and theorems developed throughout the reasoning process.\ Since summarization is inherently a lossy compression strategy, there may be steps during generation where high-level summaries fall short.\ For certain reasoning steps, accessing the full details becomes essential, whether to verify a subtle constraint, apply a specific technique, or build upon a particular intermediate result.

\subsection{Reasoning with Summarization}
By default, LLMs do not explicitly summarize their reasoning after each paragraph.\ To induce this behavior in smaller models, we fine-tune them to generate summaries after each reasoning segment.\ To achieve this, we first augment the Bespoke-17K reasoning dataset \citep{bespoke_stratos} with paragraph-level summaries.

Specifically, for each example, we segment the text within the thought token boundaries into paragraphs.\ Then, we use a larger model, Llama3-70B to summarize each segment into a concise summary.\ To create distinct, modular summaries and maintain computational efficiency during data preparation, each summary is generated based only on the prompt instruction and its corresponding paragraph.\  Summary delimiter tokens <|begin\_of\_summary|> and <|end\_of\_summary|> are inserted before a summary begins and after it ends, respectively, to explicitly mark the boundaries of each summary ((see Appendix \ref{appendix:summarization} for an example).\ We then fine-tune a base reasoning model with LoRA \citep{hu2021loralowrankadaptationlarge} to enable summary generation during inference.\ While our approach requires an initial training stage, we note that larger models like Llama3-70B, GPT4, etc.\ \citep{grattafiori2024llama3herdmodels, openai2024gpt4technicalreport}, if prompted, can generate summaries in a specific format, and the additional fine-tuning step may be unnecessary.\ Thus, the historical context is partitioned into corresponding full-text and summary segments.

\begin{figure*}[!t]
\centering
\includegraphics[width=0.9\textwidth]{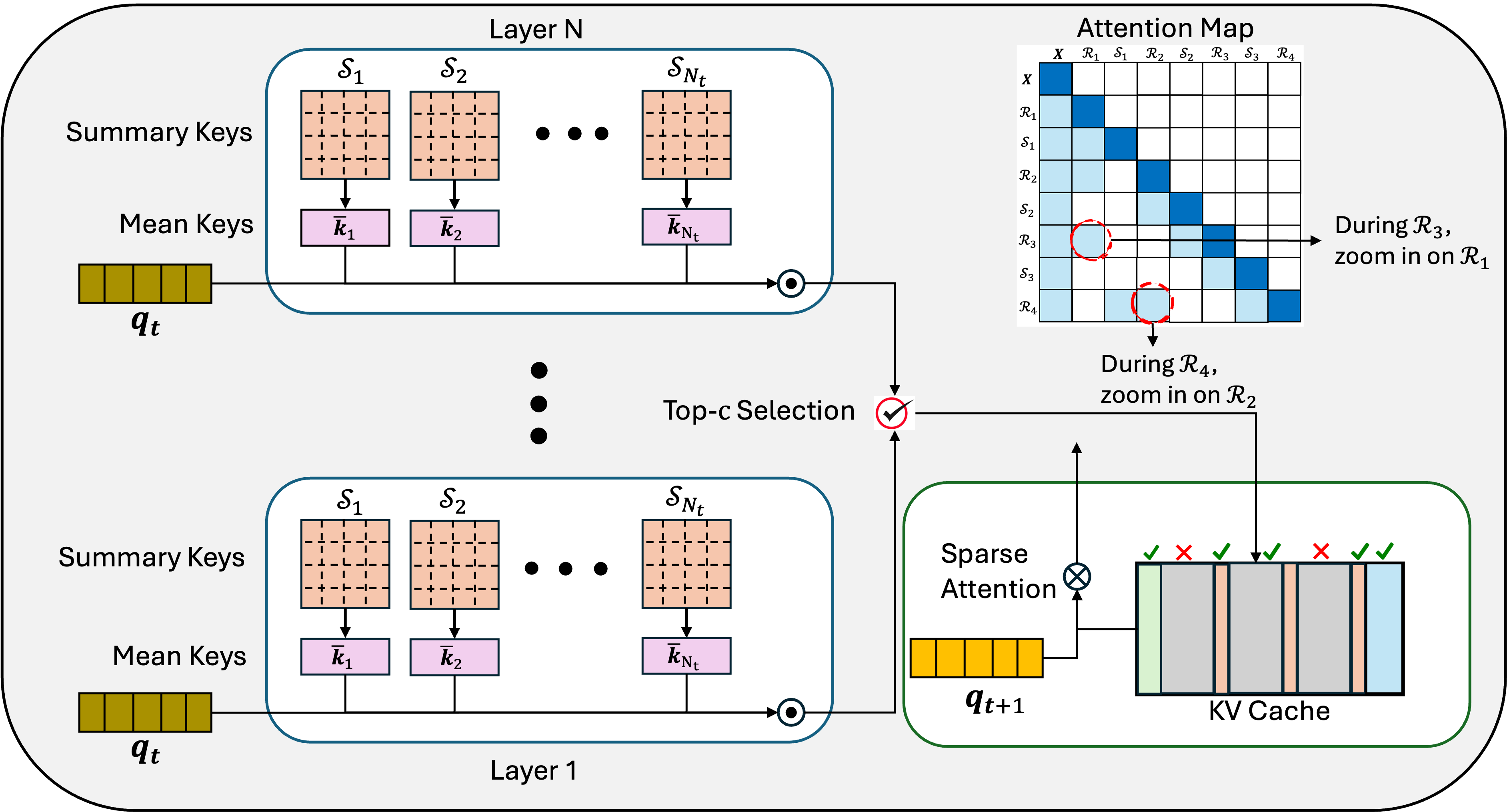}
\caption{An overview of the \kv{} workflow.
    Mean summary keys are computed and cached on GPU.\ When the KV cache selection needs to be updated, approximate attention scores are calculated across all layers. A global top-$c$ selection identifies which summaries to zoom into.\ The final sparse KV cache loaded to GPU is a multi-granularity mix of an attention sink (green), selected detailed segments, summaries, and fixed recent window (blue). The symbolic attention map illustrates the dynamic nature of ZoomR during decoding.}
\label{fig1}
\end{figure*}

\subsection{The ZoomR Algorithm}
Building upon the model's ability to generate summaries, we introduce the core algorithm of ZoomR.\ The objective is to dynamically construct a multi-granularity KV cache at each reasoning step, focusing on the most important parts of the generation context and retaining a compressed view of the rest.\ This process unfolds in four stages: representation, scoring, aggregation, and context construction (see Figure \ref{fig1}).

The output of the fine-tuned model partitions the generation context into two types of segments.\ The regular segments ($\mathcal{R}_i$) are sets of token indices corresponding to the original, detailed segments of the reasoning history and each summary segment ($\mathcal{S}_i$) is a set of token indices for the compressed summary of a regular segment, where $i$ denotes the summary number and its corresponding regular segment. The ZoomR algorithm operates by dynamically selecting a mix of these two segment types.

\paragraph{Representing Context with Mean Summary Keys.}
To efficiently score the importance of historical segments, each summary segment $\mathcal{S}_i$ is first compressed into a single representative vector for each attention head in each layer.\ This vector, termed the mean summary key, is computed by averaging the key vectors of all tokens within the summary.\ This is motivated by the observation that keys within a short, coherent text segment exhibit locality \cite{sun2025shadowkvkvcacheshadows}.\ Formally, for each summary Segment $\mathcal{S}_i$, layer $l \in \{1,\dots, N_L\}$, and head $h \in \{1, \dots, H\}$, the mean summary key is:
\begin{equation*}
\bar{\bm{k}}_i^{(l,h)} = \frac{1}{|\mathcal{S}_i|} \sum_{j \in \mathcal{S}_i} \bm{k}_j^{(l,h)},
\end{equation*}
where $\bm{k}_j^{(l,h)}$ denotes the corresponding key vector for layer $l$ and head $h$. These mean summary keys are cached on the GPU for rapid access during decoding.

\paragraph{Per-Head Importance Scoring and Voting.} At a given generation step $t$, the model's current query vector, $\bm{q}_t^{(l,h)}$, holds information about the context needed to generate the next token.\ Since query dynamics change during generation, the notion of importance and KV cache selection is greatly dependent on $\bm{q}_t^{(l,h)}$.\ We use this query to perform a per-head importance assessment.\ Specifically, for each head, we compute an approximate attention score through an inner product between the query and every mean summary key:
\begin{align*}
\alpha_i^{(l,h)} &= (\bm{q}_t^{(l,h)})^{\top} \bar{\bm{k}}_i^{(l,h)}.
\end{align*}
Each attention head then ``votes'' for the summaries it deems most important by selecting the indices of the top-$k$ highest-scoring summaries,
\begin{align*}
\mathcal{I}_{\text{top-k}}^{(l,h)} &= \underset{i \in \{1, \dots, N_t\}}{\arg\text{top-k}}(\alpha_i^{(l,h)}),
\end{align*}
where $N_t$ is the number of summaries generated up until time $t$. This initial filtering identifies a candidate set of important segments from the perspective of each head.

\paragraph{Global Aggregation and Consensus.}
To form a global decision, the individual votes from all heads and layers are aggregated.\ We first collect the union of all unique summary indices selected across heads and layers, thereby forming a global candidate set:
{\setlength{\abovedisplayskip}{4pt}
\setlength{\belowdisplayskip}{4pt}
\begin{align*}
\mathcal{I}_{\text{all}} &= \bigcup_{l=1}^{N_{L}} \bigcup_{h=1}^{H} \mathcal{I}_{\text{top-k}}^{(l,h)}. 
\end{align*}
Next, we count the total number of votes each summary in this candidate set received:
\begin{align*}
    v_i &= \sum_{l=1}^{N_{L}} \sum_{h=1}^{H} \mathbb{I}(i \in \mathcal{I}_{\text{top-k}}^{(l,h)}),
\end{align*}}
where $\mathbb{I}(\cdot)$ is the indicator function.\ The $c$ summaries with the highest vote counts are designated as the consensus set, $\mathcal{I}_{c}$.\ These represent segments identified as highly important by a majority of attention heads and layers.

\paragraph{Constructing the Final Multi-Granularity Context.}
The final KV cache is constructed by combining context at different resolutions based on the consensus results.\ Specifically, for summaries in the consensus set $\mathcal{I}_{c}$, we ``zoom in'' by including their corresponding original, full-text segments ($\mathcal{R}_i$).\ This ensure that segments consistently identified as important across heads and layers are restored  to full resolution, preserving fine-grained information critical for accurate intermediate reasoning while maintaining compression elsewhere. For the remaining summaries in the global candidate set, defined as $\mathcal{I}_s = \mathcal{I}_{\text{all}} \setminus \mathcal{I}_{c}$, we retain their compressed summary representations, $\mathcal{S}_i$.\ These segments are deemed relevant by at least one head but lack model-wide consensus.

To maintain fundamental coherence, this dynamically selected context is augmented with two critical, static components: the initial prompt tokens ($\mathcal{I}_p$), which act as an attention sink, and a sliding window of the most recently generated tokens ($\mathcal{I}_w$).\ The final set of token indices, $\mathcal{I}_f$, for the reduced KV cache is therefore obtained using the union of these components:
\begin{align*}
    \mathcal{I}_{f} = \mathcal{I}_p \cup \mathcal{I}_w \cup \left(\bigcup_{i \in \mathcal{I}_{c}} \mathcal{R}_i\right) \cup \left(\bigcup_{i \in \mathcal{I}_{s}} \mathcal{S}_i\right).
\end{align*}

At the subsequent generation step, the attention mechanism will compute over the keys and values corresponding only to the indices in $\mathcal{I}_f$.\ This dynamic, global selection process allows the model to form a holistic view of context relevance, enabling it to flexibly adjust its attentional focus between high-level summaries and fine-grained details as needed depending on the current query.\ The full algorithm for ZoomR can be found in Appendix \ref{appendix_alg}.

\begin{table*}[t]
    \caption{Performance comparison of Llama and Qwen models across GPQA Diamond, AIME2025, and MATH500 benchmarks using various approaches. Results are reported in percent (\%).}
    \centering
    \begin{adjustbox}{max width=0.7\textwidth}
    \begin{tabular}{l|ccccc}
    \toprule
    Model & Approach & GPQA Diamond & AIME2025 & MATH500 & Average \\
    \midrule
    \multirow{5}{*}{Llama} & Vanilla & 48.9 & 26.7 & 88.9 & 54.8 \\
    \cmidrule{2-6}
    & StreamingLLM & 34.7 & 13.3 & 74.7 & 40.9 \\
    & H2O & 36.8 & 16.7 & 76.5 & 43.3 \\
    & SumR & 37.3 & 16.7 & 77.8 & 43.9 \\
    & \textbf{ZoomR} & \textbf{43.4} & \textbf{26.7} & \textbf{83.0} & \textbf{51.0} \\
    \midrule
    \multirow{5}{*}{Qwen} & Vanilla & 49.0 & 30.0 & 92.5 & 57.2 \\
    \cmidrule{2-6}
    & StreamingLLM & 37.9 & 13.3 & 77.9 & 43.0 \\
    & H2O & 39.1 & 16.7 & 79.4 & 45.1 \\
    & SumR & 37.6 & 13.3 & 80.1 & 43.7 \\
    & \textbf{ZoomR} & \textbf{42.8} & \textbf{23.3} & \textbf{87.8} & \textbf{51.3} \\
    \bottomrule
    \end{tabular}
    \end{adjustbox}
    \label{commonsense}
\end{table*}

\subsection{System Implementation and Inference}
The practical application of ZoomR involves a memory management strategy that dynamically moves KV cache data between high-capacity CPU memory and fast GPU memory.\ The process is divided into a one-time prefill stage and a repeated decoding stage.
\paragraph{Prefill Stage.} During the prefill stage, the input prompt $\bm{X}$ of length $N_p$ is processed to generate the initial KV cache.\ To conserve GPU memory, we offload these initial KVs to CPU memory.\ Prompt tokens are crucial to maintain coherent LLM inference \cite{xiao2024efficientstreaminglanguagemodels}, and this is accentuated in complex reasoning tasks.\ Thus, we choose to retain the full KV cache from the prefill stage.\ In many mathematical reasoning benchmarks, $N_p \ll N_g$, so the relative cost is minimal.

\paragraph{Decoding Stage.} To optimize for memory efficiency, at each generation step, only the subset of the KV cache specified by the index set $\mathcal{I}_f$ is loaded from CPU to GPU.\ After the attention computation for the current token is complete, the newly generated KV pair is appended to the full cache on the CPU.\ While this CPU-GPU data transfer introduces latency dependent on Peripheral Component Interconnect Express (PCIe) bandwidth, it enables reasoning over contexts far exceeding the available GPU memory.

To optimize performance, we introduce two key efficiency improvements.\ First, recomputing the selection set $\mathcal{I}_f$ at every step is computationally expensive.\ We only update the selection set at semantic boundaries, such as end-of-sentence, amortizing the cost of selection over multiple decoding steps (see Appendix \ref{app:sem_bound} for details).\ Second, to minimize peak GPU memory usage and hide data transfer latency, we implement a pipelined, layer-by-layer execution strategy.\ For each layer of the model, the required KV cache slices are transferred to the GPU.\ Once the layer's computation is finished, its KVs are asynchronously transferred back to the CPU while the KVs for the next layer are prefetched.\ This overlaps data transfer with computation, reducing GPU idle time.

\subsection{Theoretical Memory Analysis}

ZoomR's efficiency comes from only loading a small subset of the KV cache from CPU to GPU at each decoding step.\ We quantify this by comparing the memory footprint of a full KV cache with that of ZoomR's active sparse cache on the GPU.

The KV cache size selected by ZoomR corresponds to the number of tokens in the selection set $\mathcal{I}_f$.\ Let us consider each component in $\mathcal{I}_f$. Define the number of prompt tokens as $N_p$, the consensus selected $c$ regular segments of average length $\bar{L}_R$, $(|\mathcal{I}_{\text{all}}| - c)$ summary segments of average length $\bar{L}_S$, and size of the recent window as $N_w$.\ In the worst case, the number of selected summaries $|\mathcal{I}_{\text{all}}|$ is bounded by the total number of summaries generated during decoding.\ During generation, the full KV cache stores all previously generated tokens, reaching a size of $N_g$ at the final step.\ Thus, the memory savings (MS) can be computed as:
\begin{equation*}
\text{MS} = \frac{N_p + N_g}{N_{p} + c \cdot \bar{L}_R + (|\mathcal{I}_{\text{all}}| - c) \cdot \bar{L}_S + N_w}
\end{equation*}

For $N_p=512$, $N_g=16384$, $N_w=512$, $c=2$, $\bar{L}_R=250$, $\bar{L}_S=20$, and $|\mathcal{I}_{\text{all}}| = 80$, the memory savings is $5.48\times$.

Crucially, the denominator grows sub-linearly with the sequence length as the number of summaries increases at a smaller rate.\ This ensures that memory savings improve as the generation length increases.\ For instance, doubling $N_g$ to 32K yields savings of approximately $7.11\times$.

\section{Experiments}
In this section we present the experiment setup and evaluate ZoomR using two reasoning models: Qwen2.5-7B and Llama-3.1-8B.\ For each model, we perform supervised fine-tuning with a max context length of 16K tokens using the augmented Bespoke-17K dataset.\ We refer to these fine-tuned models as the vanilla baselines.\ Specifically, we use the distilled reasoning models from R1-Distill, Deepseek-R1-Distill-Qwen2-7B and Deepseek-R1-Distill-Llama3-8b \citep{deepseekai2025deepseekr1incentivizingreasoningcapability} as the base models for fine-tuning.

We compare ZoomR against three other baseline approaches: StreamingLLM, H2O, and SumR. StreamingLLM uses a static KV cache selection policy based on fixed attention sinks and a recent sliding window, while H2O uses a dynamic approach that selects tokens based on importance scores at each generation step.\ To ensure a fair comparison between ZoomR, StreamingLLM, and H2O, we first compute the GPU budget used by ZoomR for each task, and then set the equivalent budget for StreamingLLM and H2O.\ We also introduce SumR, a simplified variant of ZoomR that retains only the summary tokens and does not perform any detailed token retrieval. We evaluate all methods on both math and reasoning tasks, including MATH500, GPQA Diamond, and AIME2025.\ All experiments are run on NVIDIA H100 GPUs.

\begin{figure*}[t]
\centering

\begin{subfigure}[b]{0.32\textwidth}
    \centering
    \includegraphics[width=\textwidth]{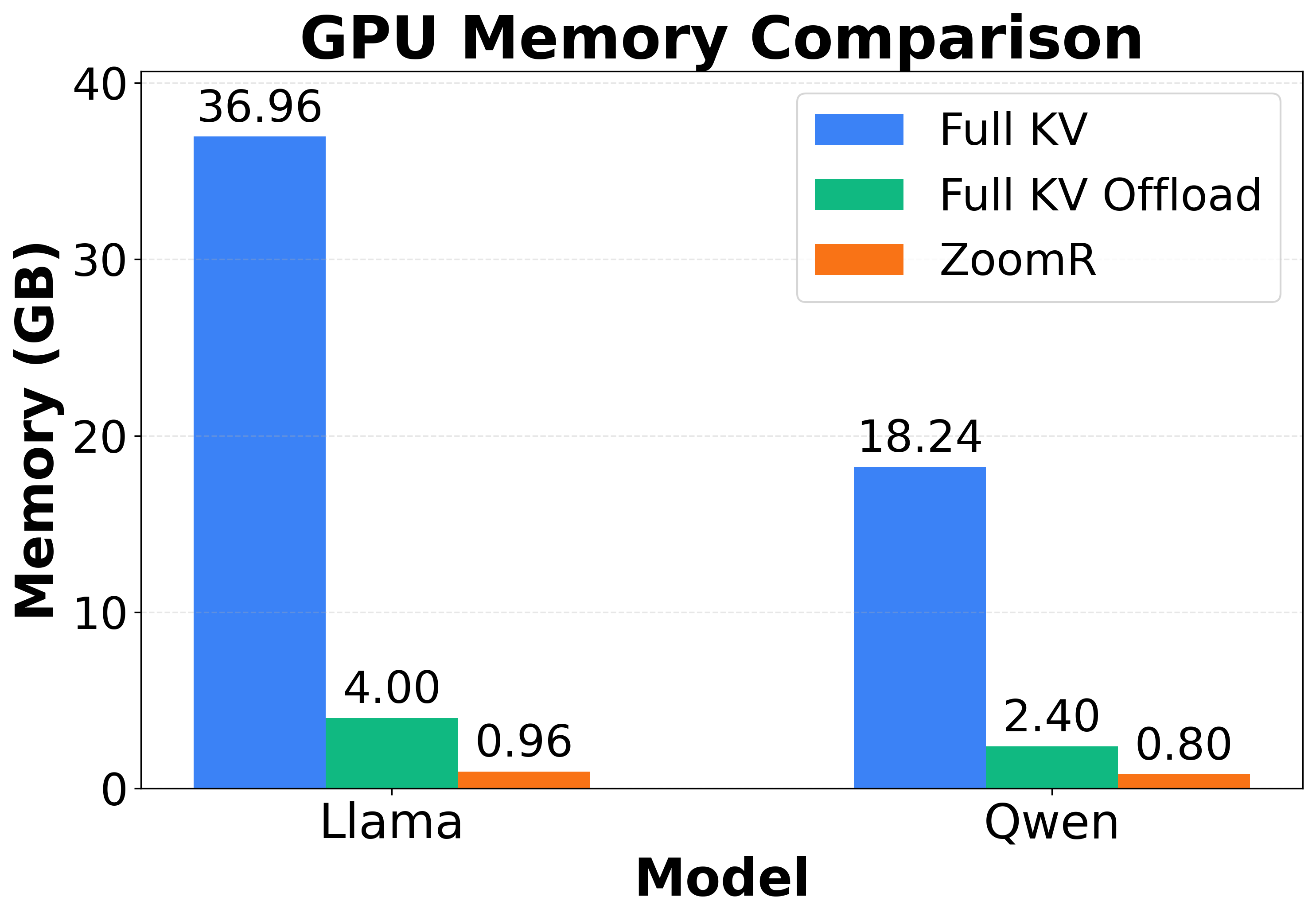}
    \caption{KV cache comparison between ZoomR and full KV cache.}
    \label{fig:kv_cache}
\end{subfigure}
\hfill
\begin{subfigure}[b]{0.32\textwidth}
    \centering
    \includegraphics[width=\textwidth]{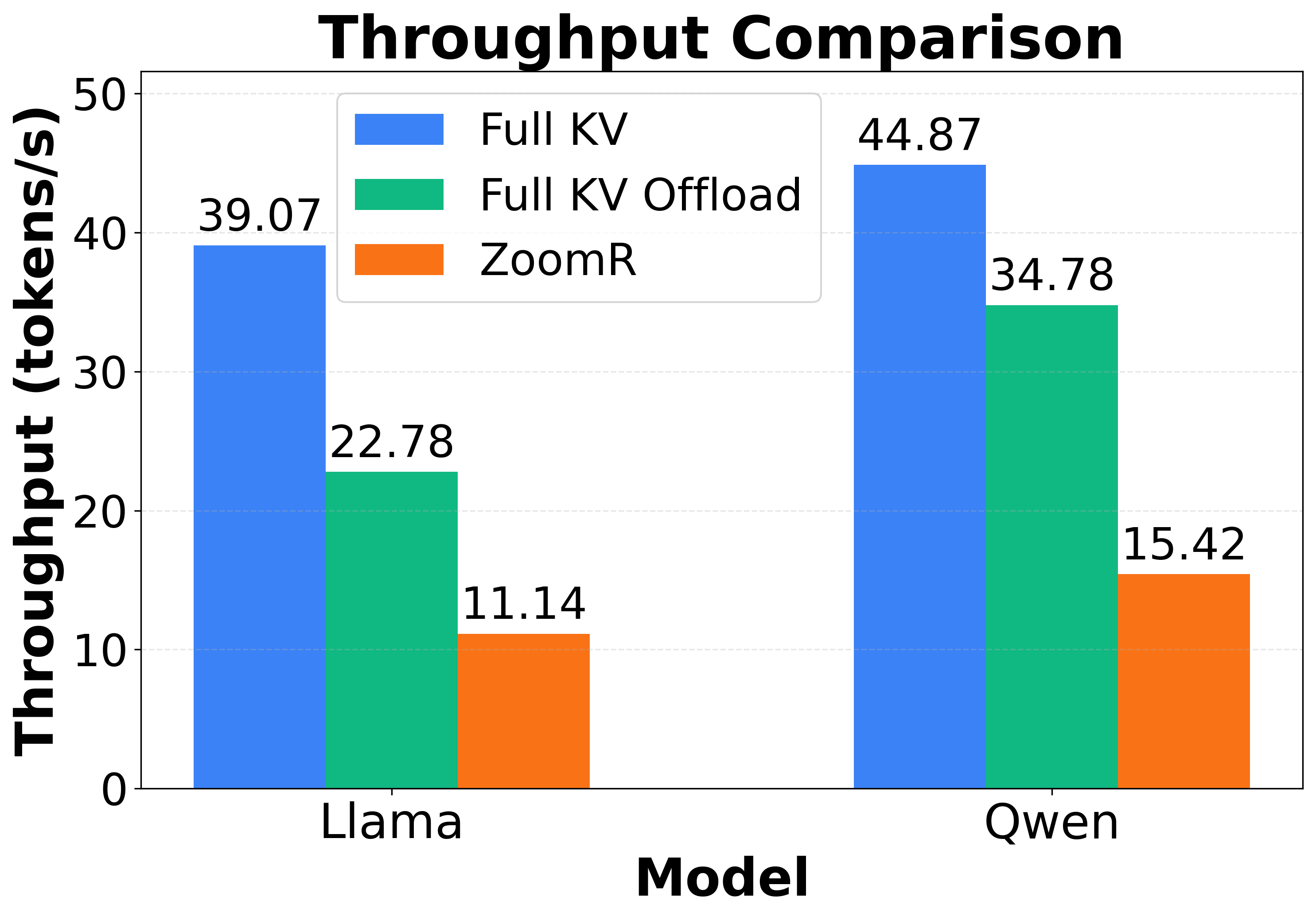}
    \caption{Runtime comparison between ZoomR and full KV cache.}
    \label{fig:runtime}
\end{subfigure}
\hfill
\begin{subfigure}[b]{0.32\textwidth}
    \centering
    \includegraphics[width=\textwidth]{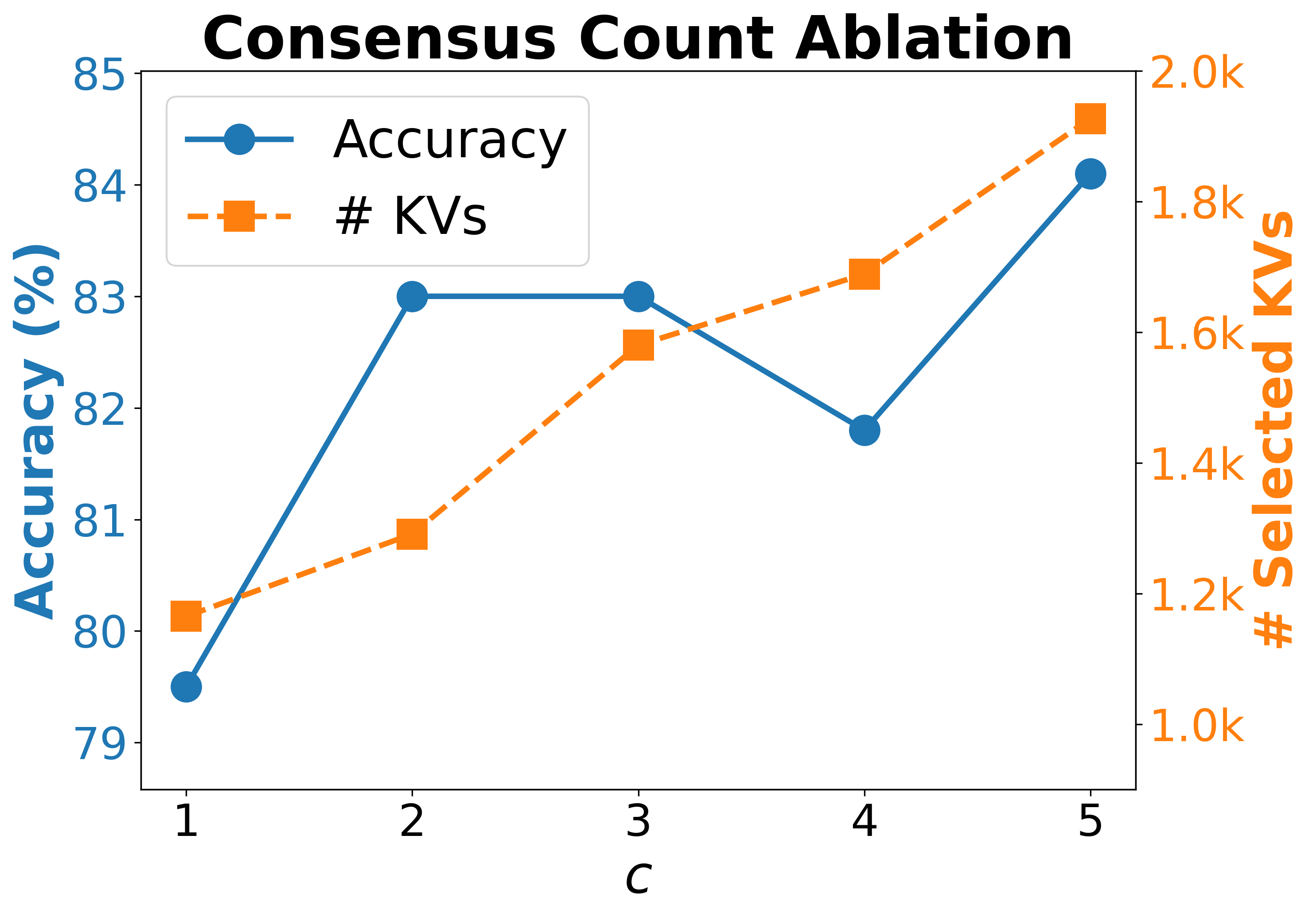}
    \caption{Impact of $c$ on accuracy and number of KVs selected.}
    \label{fig:impact_of_c}
\end{subfigure}

\vspace{1em} 

\begin{subfigure}[b]{0.32\textwidth}
    \centering
    \includegraphics[width=\textwidth]{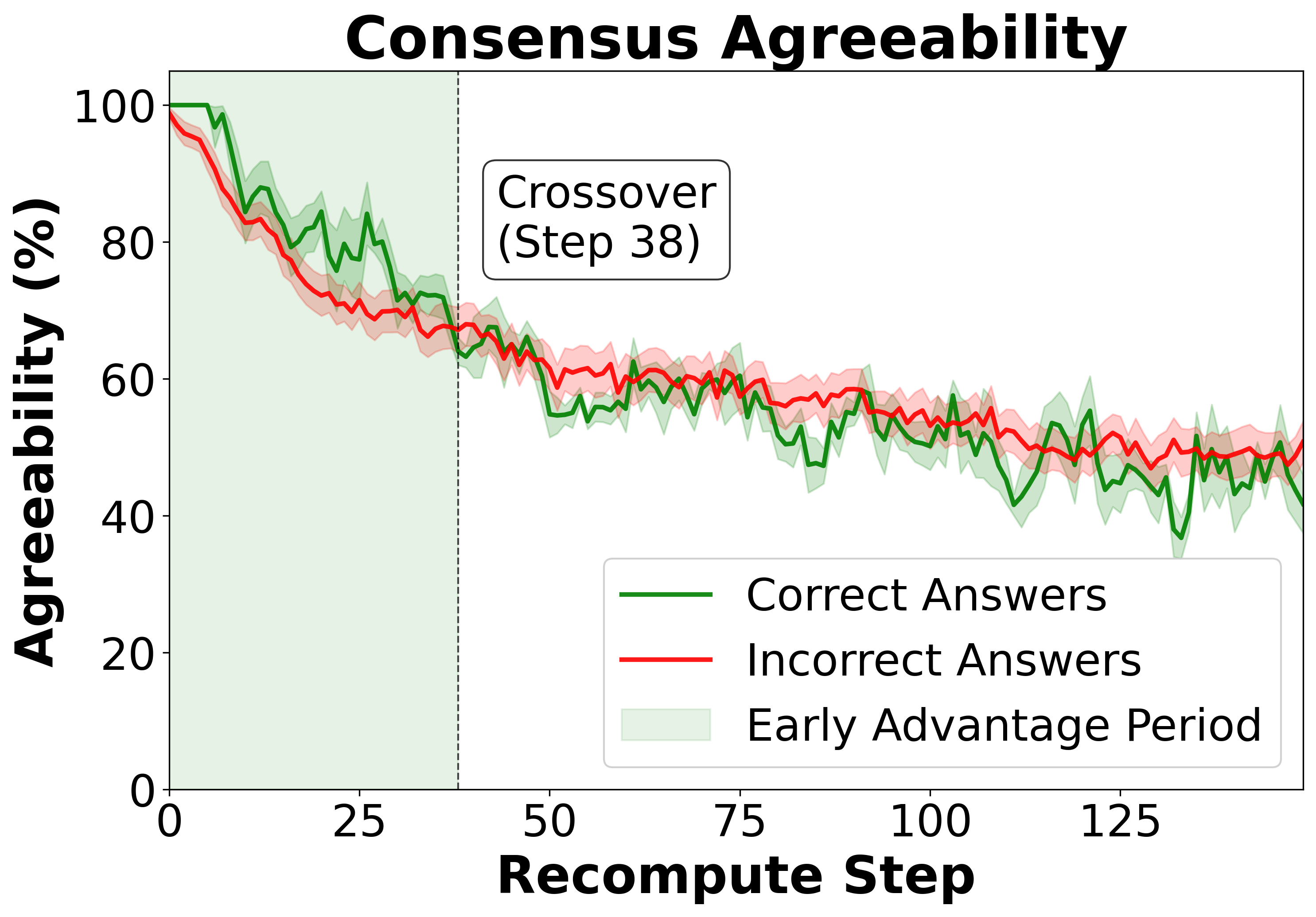}
    \caption{Aggregate agreeability over time for correct and incorrect answers.}
    \label{fig:aggregate_agreeability}
\end{subfigure}
\hfill
\begin{subfigure}[b]{0.32\textwidth}
    \centering
    \includegraphics[width=\textwidth]{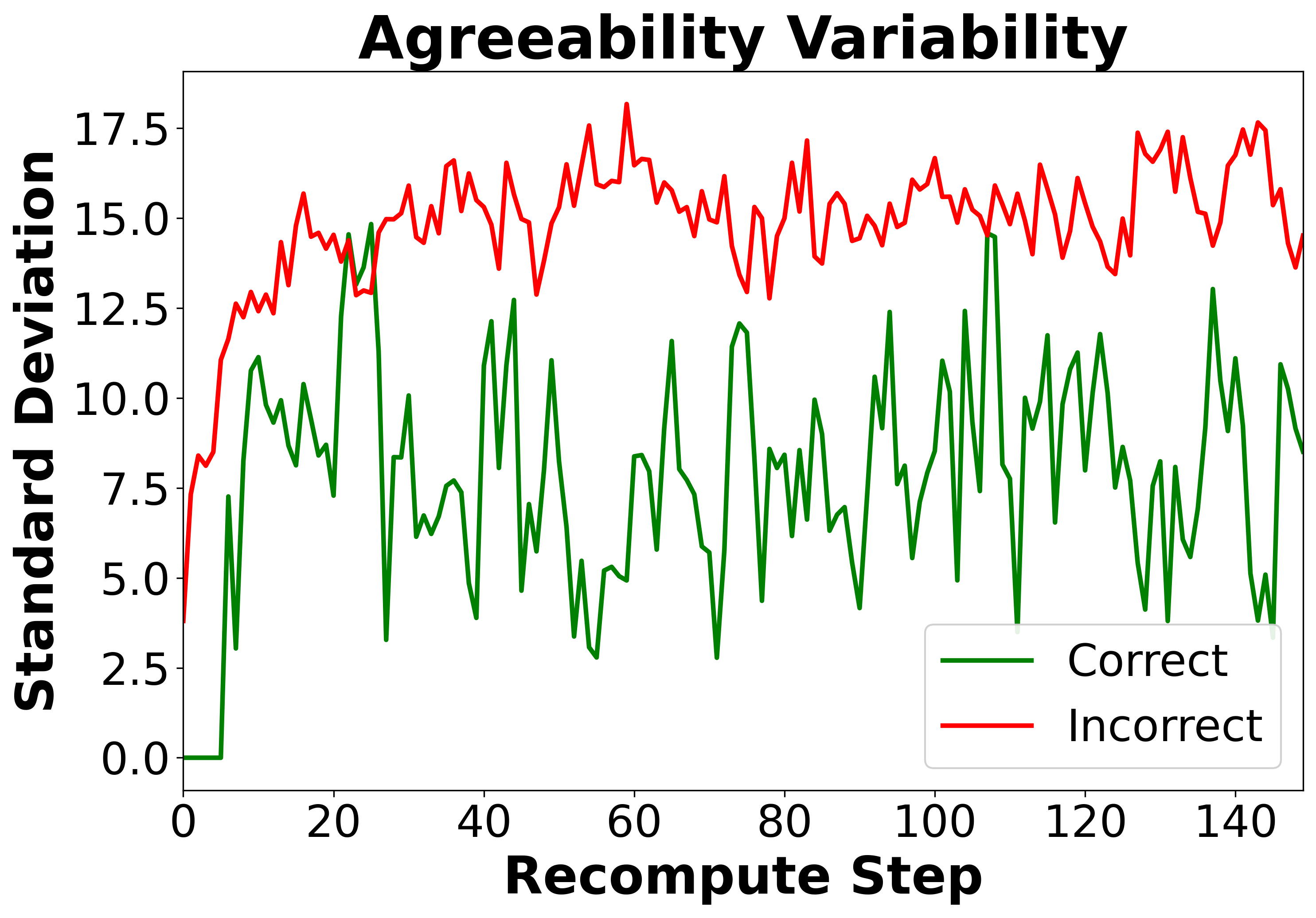}
    \caption{Agreeability standard deviation over time for correct and incorrect answers.}
    \label{fig:agreeability_variability}
\end{subfigure}
\hfill
\begin{subfigure}[b]{0.32\textwidth}
    \centering
    \includegraphics[width=\textwidth]{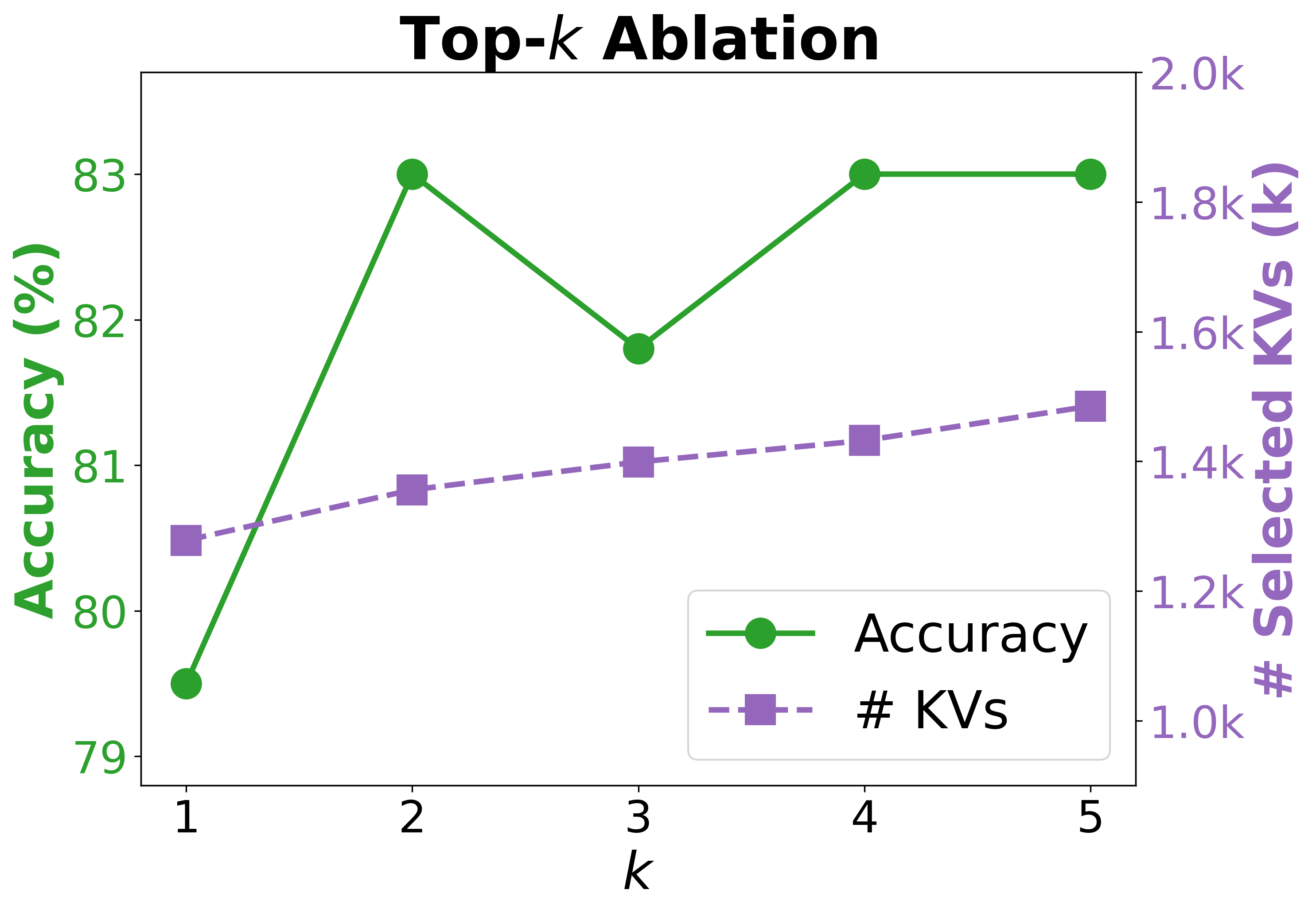}
    \caption{Impact of $k$ on accuracy and number of KVs selected.}
    \label{fig:impact_of_k}
\end{subfigure}

\caption{Efficiency, consensus, and ablation studies.\ Fig.\ (a,b) shows the GPU memory and throughput during inference, respectively.\ Fig.\ (d,e) show consensus dynamics between correct and incorrect answers over time.\ Fig.\ (c,f) illustrates the relationships between $c$ and $k$ on performance and KV cache selection size.}
\label{fig:efficiency_comparison_full}
\end{figure*}

\paragraph{Results.}
As shown in Table \ref{commonsense}, ZoomR significantly outperforms baseline approaches for all tasks, achieving accuracy close to the full KV cache.\ For the challenging AIME2025 benchmark, ZoomR is particularly effective, matching the performance of the full KV cache with the Llama model while remaining highly competitive using Qwen.\ Notably, these results demonstrate the benefits of dynamically zooming in to detailed context.\ ZoomR sees an $8\%$ average accuracy improvement over SumR, indicating that only selecting summaries leads to loss of information, thereby impacting performance.

ZoomR is significantly more memory efficient than the vanilla baselines.\ We compare against both cases of full KV cache on GPU and offloaded to CPU.\ Figure \ref {fig:kv_cache} shows the GPU memory utilization during inference generation in these cases using a 16K generation length and batch size of 16.\ Note that we exclude the model weights' contribution to memory. ZoomR saves more than 20$\times$ GPU memory for both Llama and Qwen based models compared to using the full KV cache stored on GPU, and more than $4\times$ when full KV cache is offloaded to CPU.\ Since both the average summary lengths and consensus count $c$ are fixed, the savings increase significantly with longer sequence length and larger batch sizes.\ As a tradeoff, due to the CPU-GPU data transfer latency, ZoomR incurs a reduction in throughput (see Figure \ref{fig:runtime}), which is limited by the PCIe bandwidth.\ Therefore, ZoomR provides a major memory efficiency boost at the cost of reduced throughput, while maintaining strong performance compared to other KV cache compression methods.

\paragraph{Consensus Dynamics in Thought Compression.} Our analysis reveals that consensus patterns between attention heads serve as a useful diagnostic signal for reasoning quality in mathematical tasks.\ We define an agreeability metric $AG$, as a measure of consensus among all attention heads regarding the importance of summary segments.\ Specificially, it quantifies the fraction of total ``votes'' that are assigned to the most frequently chosen summaries, i.e.\ the consensus set, $AG = \sum_{i \in \mathcal{I}_{c}}v_i / \sum_{j \in \mathcal{I}_{\text{all}}} v_j$.
We observe a striking early advantage for correct answers, with 5.7\% higher agreeability in the first 38 recompute steps compared to incorrect answers (see Figure \ref{fig:aggregate_agreeability}).\ This early consensus advantage suggests that correct reasoning converges quickly to a coherent solution path, while incorrect reasoning exhibits more exploratory behavior. The temporal dynamics show a clear crossover at step 38, after which incorrect answers demonstrate higher agreeability than correct ones. This pattern indicates that incorrect answers may require more extensive exploration or get stuck on reasoning paths, unable to generate the correct answer.

Critically, incorrect answers show significantly higher variability in $AG$ throughout the reasoning process, with a standard deviation of 14.82\% compared to 7.87\% for correct answers (see Figure \ref{fig:agreeability_variability}).\ This increased variability reflects the inherent instability of incorrect reasoning paths.\ This difference persists across all time windows, suggesting that consensus stability itself may be a fundamental property distinguishing correct from incorrect mathematical reasoning.

These findings support our hypothesis that ZoomR leverages consensus dynamics to guide reasoning quality through selective KV caching, offering new insights into how models can ``self-assess'' reasoning quality through internal agreement patterns during selective attention mechanisms.

\paragraph{Ablation Studies.} We conduct ablation studies to assess the impact of the hyperparameters consensus count $c$ and top-$k$, on performance.\ Specifically, we run the experiments on MATH500 using the Llama model.\ With $c=1$, only one summary is expanded to fine-details, and the performance drops by $3\%$ compared to $c=2$.\ However, for $c \ge 2$, there is only a minor improvement of $1\%$, while the number of KVs being selected increases by over $40\%$, as more summaries are being expanded (see Figure \ref{fig:impact_of_c}).\ For $k=1$, the performance is also noticeably worse, seeing close to a $3\%$ drop compared to $k=2$.\ This suggests that selecting only one top summary per head discards important context, reducing aggregation quality.\ However, when $k \ge 2$, the accuracy plateaus (see Figure \ref{fig:impact_of_k}).\ The increase in number of tokens selected is also more gradual since increasing $k$ only increases the number of summaries to select, which contribute much less to the KV cache.

\section{Related Work}
A central challenge in efficient reasoning is reducing both memory and compute overhead during inference. One line of work addresses this by reducing the number of tokens generated during inference \citep{sui2025stopoverthinkingsurveyefficient}.\ Latent reasoning and summarization based approaches enable efficient reasoning by summarizing the intermediate thinking steps. For instance, CODI and Coconut~\citep{shen2025codi, hao2024training} train an LLM to compress a natural-language chain-of-thought into a continuous latent trajectory.\ LightThinker~\citep{zhang2025lightthinkerthinkingstepbystepcompression} trains an LLM to replace verbose intermediate reasoning with concise “gist” tokens and discards the full intermediate thoughts during decoding.\ While these methods offer inference-time efficiency, they rely on latent or opaque representations that are difficult to interprety or verify.\ Moreover, memory savings are typically a secondary concern. In contrast, our focus is on memory-centric techniques, specifically those involving KV cache compression.

\paragraph{KV Cache Compression.} KV cache compression reduces both storage and the computational cost of LLM inference. KV cache quantization reduces memory by storing the key and value tensors with lower numerical precision \cite{https://doi.org/10.13140/rg.2.2.28167.37282, hooper2025kvquant10millioncontext}. In addition, KVs can also be compressed with low-rank projections along the model dimension~\citep{saxena2024eigen, zhu2025ojakvcontextawareonlinelowrank}.\ These techniques are complementary to our method, as they operate on the representation level (e.g., precision or dimensionality).

Token selection methods retain or evict tokens based on learned or heuristic policies. StreamingLLM~\citep{xiao2024efficientstreaminglanguagemodels} exploits the attention-sink effect to retain the initial tokens and the most recent context.\ SnapKV and H2O~\citep{li2024snapkv, zhang2023h2oheavyhitteroracleefficient} rank and retain important tokens using scoring mechanisms.\ Quest, ShadowKV, and SentenceKV~\citep{tang2024questqueryawaresparsityefficient, sun2025shadowkvkvcacheshadows, zhu2025sentencekvefficientllminference} move the KV cache generated during the prefill stage from GPU to CPU memory and retrieve a subset of KVs to attend to during decoding.\ While there is a growing literature of KV cache selection methods that enable memory efficiency for long context input, most of these approaches assume that the full KV cache is on GPU memory during the decoding stage, and are not designed for long-output reasoning tasks. Our work addresses this overlook scenario.

\section{Conclusion}
In this paper, we present ZoomR, a dynamic KV cache selection policy that enables memory efficient reasoning.\ By fine-tuning LLMs to generate summaries, and dynamically selecting summaries to ``zoom'' into based on a global consensus between attention heads, ZoomR enables efficient reasoning with higher fidelity.\ Extensive experiments show that ZoomR consistently outperforms other baseline approaches, maintaining competitive performance compared to full KV cache.

\section*{Limitations}

One key factor that can influence performance is quality of summaries used during data augmentation.\ In this work, we use summaries generated by the Llama-70B model, and do not evaluate the effect of using summaries generated by larger, more powerful models.\ The granularity of segment selection for summarization is fixed at the paragraph level. It remains unclear whether summarizing at finer or coarser granularities would improve or degrade ZoomR's performance. Finally, ZoomR is primarily evaluated on mathematical reasoning tasks. Future work should explore its applicability to other domains such as code generation and creative writing, where reasoning structures may differ significantly.


\bibliography{custom}

@misc{deepseekai2025deepseekr1incentivizingreasoningcapability,
      title={DeepSeek-R1: Incentivizing Reasoning Capability in LLMs via Reinforcement Learning}, 
      author={DeepSeek-AI and Daya Guo and Dejian Yang and Haowei Zhang and Junxiao Song and Ruoyu Zhang and Runxin Xu and Qihao Zhu and Shirong Ma and Peiyi Wang and Xiao Bi and Xiaokang Zhang and Xingkai Yu and Yu Wu and Z. F. Wu and Zhibin Gou and Zhihong Shao and Zhuoshu Li and Ziyi Gao and Aixin Liu and Bing Xue and Bingxuan Wang and Bochao Wu and Bei Feng and Chengda Lu and Chenggang Zhao and Chengqi Deng and Chenyu Zhang and Chong Ruan and Damai Dai and Deli Chen and Dongjie Ji and Erhang Li and Fangyun Lin and Fucong Dai and Fuli Luo and Guangbo Hao and Guanting Chen and Guowei Li and H. Zhang and Han Bao and Hanwei Xu and Haocheng Wang and Honghui Ding and Huajian Xin and Huazuo Gao and Hui Qu and Hui Li and Jianzhong Guo and Jiashi Li and Jiawei Wang and Jingchang Chen and Jingyang Yuan and Junjie Qiu and Junlong Li and J. L. Cai and Jiaqi Ni and Jian Liang and Jin Chen and Kai Dong and Kai Hu and Kaige Gao and Kang Guan and Kexin Huang and Kuai Yu and Lean Wang and Lecong Zhang and Liang Zhao and Litong Wang and Liyue Zhang and Lei Xu and Leyi Xia and Mingchuan Zhang and Minghua Zhang and Minghui Tang and Meng Li and Miaojun Wang and Mingming Li and Ning Tian and Panpan Huang and Peng Zhang and Qiancheng Wang and Qinyu Chen and Qiushi Du and Ruiqi Ge and Ruisong Zhang and Ruizhe Pan and Runji Wang and R. J. Chen and R. L. Jin and Ruyi Chen and Shanghao Lu and Shangyan Zhou and Shanhuang Chen and Shengfeng Ye and Shiyu Wang and Shuiping Yu and Shunfeng Zhou and Shuting Pan and S. S. Li and Shuang Zhou and Shaoqing Wu and Shengfeng Ye and Tao Yun and Tian Pei and Tianyu Sun and T. Wang and Wangding Zeng and Wanjia Zhao and Wen Liu and Wenfeng Liang and Wenjun Gao and Wenqin Yu and Wentao Zhang and W. L. Xiao and Wei An and Xiaodong Liu and Xiaohan Wang and Xiaokang Chen and Xiaotao Nie and Xin Cheng and Xin Liu and Xin Xie and Xingchao Liu and Xinyu Yang and Xinyuan Li and Xuecheng Su and Xuheng Lin and X. Q. Li and Xiangyue Jin and Xiaojin Shen and Xiaosha Chen and Xiaowen Sun and Xiaoxiang Wang and Xinnan Song and Xinyi Zhou and Xianzu Wang and Xinxia Shan and Y. K. Li and Y. Q. Wang and Y. X. Wei and Yang Zhang and Yanhong Xu and Yao Li and Yao Zhao and Yaofeng Sun and Yaohui Wang and Yi Yu and Yichao Zhang and Yifan Shi and Yiliang Xiong and Ying He and Yishi Piao and Yisong Wang and Yixuan Tan and Yiyang Ma and Yiyuan Liu and Yongqiang Guo and Yuan Ou and Yuduan Wang and Yue Gong and Yuheng Zou and Yujia He and Yunfan Xiong and Yuxiang Luo and Yuxiang You and Yuxuan Liu and Yuyang Zhou and Y. X. Zhu and Yanhong Xu and Yanping Huang and Yaohui Li and Yi Zheng and Yuchen Zhu and Yunxian Ma and Ying Tang and Yukun Zha and Yuting Yan and Z. Z. Ren and Zehui Ren and Zhangli Sha and Zhe Fu and Zhean Xu and Zhenda Xie and Zhengyan Zhang and Zhewen Hao and Zhicheng Ma and Zhigang Yan and Zhiyu Wu and Zihui Gu and Zijia Zhu and Zijun Liu and Zilin Li and Ziwei Xie and Ziyang Song and Zizheng Pan and Zhen Huang and Zhipeng Xu and Zhongyu Zhang and Zhen Zhang},
      year={2025},
      eprint={2501.12948},
      archivePrefix={arXiv},
      primaryClass={cs.CL},
      url={https://arxiv.org/abs/2501.12948}, 
}

@misc{openai2024openaio1card,
      title={OpenAI o1 System Card}, 
      author={OpenAI and : and Aaron Jaech and Adam Kalai and Adam Lerer and Adam Richardson and Ahmed El-Kishky and Aiden Low and Alec Helyar and Aleksander Madry and Alex Beutel and Alex Carney and Alex Iftimie and Alex Karpenko and Alex Tachard Passos and Alexander Neitz and Alexander Prokofiev and Alexander Wei and Allison Tam and Ally Bennett and Ananya Kumar and Andre Saraiva and Andrea Vallone and Andrew Duberstein and Andrew Kondrich and Andrey Mishchenko and Andy Applebaum and Angela Jiang and Ashvin Nair and Barret Zoph and Behrooz Ghorbani and Ben Rossen and Benjamin Sokolowsky and Boaz Barak and Bob McGrew and Borys Minaiev and Botao Hao and Bowen Baker and Brandon Houghton and Brandon McKinzie and Brydon Eastman and Camillo Lugaresi and Cary Bassin and Cary Hudson and Chak Ming Li and Charles de Bourcy and Chelsea Voss and Chen Shen and Chong Zhang and Chris Koch and Chris Orsinger and Christopher Hesse and Claudia Fischer and Clive Chan and Dan Roberts and Daniel Kappler and Daniel Levy and Daniel Selsam and David Dohan and David Farhi and David Mely and David Robinson and Dimitris Tsipras and Doug Li and Dragos Oprica and Eben Freeman and Eddie Zhang and Edmund Wong and Elizabeth Proehl and Enoch Cheung and Eric Mitchell and Eric Wallace and Erik Ritter and Evan Mays and Fan Wang and Felipe Petroski Such and Filippo Raso and Florencia Leoni and Foivos Tsimpourlas and Francis Song and Fred von Lohmann and Freddie Sulit and Geoff Salmon and Giambattista Parascandolo and Gildas Chabot and Grace Zhao and Greg Brockman and Guillaume Leclerc and Hadi Salman and Haiming Bao and Hao Sheng and Hart Andrin and Hessam Bagherinezhad and Hongyu Ren and Hunter Lightman and Hyung Won Chung and Ian Kivlichan and Ian O'Connell and Ian Osband and Ignasi Clavera Gilaberte and Ilge Akkaya and Ilya Kostrikov and Ilya Sutskever and Irina Kofman and Jakub Pachocki and James Lennon and Jason Wei and Jean Harb and Jerry Twore and Jiacheng Feng and Jiahui Yu and Jiayi Weng and Jie Tang and Jieqi Yu and Joaquin Quiñonero Candela and Joe Palermo and Joel Parish and Johannes Heidecke and John Hallman and John Rizzo and Jonathan Gordon and Jonathan Uesato and Jonathan Ward and Joost Huizinga and Julie Wang and Kai Chen and Kai Xiao and Karan Singhal and Karina Nguyen and Karl Cobbe and Katy Shi and Kayla Wood and Kendra Rimbach and Keren Gu-Lemberg and Kevin Liu and Kevin Lu and Kevin Stone and Kevin Yu and Lama Ahmad and Lauren Yang and Leo Liu and Leon Maksin and Leyton Ho and Liam Fedus and Lilian Weng and Linden Li and Lindsay McCallum and Lindsey Held and Lorenz Kuhn and Lukas Kondraciuk and Lukasz Kaiser and Luke Metz and Madelaine Boyd and Maja Trebacz and Manas Joglekar and Mark Chen and Marko Tintor and Mason Meyer and Matt Jones and Matt Kaufer and Max Schwarzer and Meghan Shah and Mehmet Yatbaz and Melody Y. Guan and Mengyuan Xu and Mengyuan Yan and Mia Glaese and Mianna Chen and Michael Lampe and Michael Malek and Michele Wang and Michelle Fradin and Mike McClay and Mikhail Pavlov and Miles Wang and Mingxuan Wang and Mira Murati and Mo Bavarian and Mostafa Rohaninejad and Nat McAleese and Neil Chowdhury and Neil Chowdhury and Nick Ryder and Nikolas Tezak and Noam Brown and Ofir Nachum and Oleg Boiko and Oleg Murk and Olivia Watkins and Patrick Chao and Paul Ashbourne and Pavel Izmailov and Peter Zhokhov and Rachel Dias and Rahul Arora and Randall Lin and Rapha Gontijo Lopes and Raz Gaon and Reah Miyara and Reimar Leike and Renny Hwang and Rhythm Garg and Robin Brown and Roshan James and Rui Shu and Ryan Cheu and Ryan Greene and Saachi Jain and Sam Altman and Sam Toizer and Sam Toyer and Samuel Miserendino and Sandhini Agarwal and Santiago Hernandez and Sasha Baker and Scott McKinney and Scottie Yan and Shengjia Zhao and Shengli Hu and Shibani Santurkar and Shraman Ray Chaudhuri and Shuyuan Zhang and Siyuan Fu and Spencer Papay and Steph Lin and Suchir Balaji and Suvansh Sanjeev and Szymon Sidor and Tal Broda and Aidan Clark and Tao Wang and Taylor Gordon and Ted Sanders and Tejal Patwardhan and Thibault Sottiaux and Thomas Degry and Thomas Dimson and Tianhao Zheng and Timur Garipov and Tom Stasi and Trapit Bansal and Trevor Creech and Troy Peterson and Tyna Eloundou and Valerie Qi and Vineet Kosaraju and Vinnie Monaco and Vitchyr Pong and Vlad Fomenko and Weiyi Zheng and Wenda Zhou and Wes McCabe and Wojciech Zaremba and Yann Dubois and Yinghai Lu and Yining Chen and Young Cha and Yu Bai and Yuchen He and Yuchen Zhang and Yunyun Wang and Zheng Shao and Zhuohan Li},
      year={2024},
      eprint={2412.16720},
      archivePrefix={arXiv},
      primaryClass={cs.AI},
      url={https://arxiv.org/abs/2412.16720}, 
}

@misc{sui2025stopoverthinkingsurveyefficient,
      title={Stop Overthinking: A Survey on Efficient Reasoning for Large Language Models}, 
      author={Yang Sui and Yu-Neng Chuang and Guanchu Wang and Jiamu Zhang and Tianyi Zhang and Jiayi Yuan and Hongyi Liu and Andrew Wen and Shaochen Zhong and Na Zou and Hanjie Chen and Xia Hu},
      year={2025},
      eprint={2503.16419},
      archivePrefix={arXiv},
      primaryClass={cs.CL},
      url={https://arxiv.org/abs/2503.16419}, 
}

@misc{hao2024traininglargelanguagemodels,
      title={Training Large Language Models to Reason in a Continuous Latent Space}, 
      author={Shibo Hao and Sainbayar Sukhbaatar and DiJia Su and Xian Li and Zhiting Hu and Jason Weston and Yuandong Tian},
      year={2024},
      eprint={2412.06769},
      archivePrefix={arXiv},
      primaryClass={cs.CL},
      url={https://arxiv.org/abs/2412.06769}, 
}

@misc{shen2025efficientreasoninghiddenthinking,
      title={Efficient Reasoning with Hidden Thinking}, 
      author={Xuan Shen and Yizhou Wang and Xiangxi Shi and Yanzhi Wang and Pu Zhao and Jiuxiang Gu},
      year={2025},
      eprint={2501.19201},
      archivePrefix={arXiv},
      primaryClass={cs.CL},
      url={https://arxiv.org/abs/2501.19201}, 
}

@misc{xia2025tokenskipcontrollablechainofthoughtcompression,
      title={TokenSkip: Controllable Chain-of-Thought Compression in LLMs}, 
      author={Heming Xia and Chak Tou Leong and Wenjie Wang and Yongqi Li and Wenjie Li},
      year={2025},
      eprint={2502.12067},
      archivePrefix={arXiv},
      primaryClass={cs.CL},
      url={https://arxiv.org/abs/2502.12067}, 
}

@misc{tang2024questqueryawaresparsityefficient,
      title={Quest: Query-Aware Sparsity for Efficient Long-Context LLM Inference}, 
      author={Jiaming Tang and Yilong Zhao and Kan Zhu and Guangxuan Xiao and Baris Kasikci and Song Han},
      year={2024},
      eprint={2406.10774},
      archivePrefix={arXiv},
      primaryClass={cs.CL},
      url={https://arxiv.org/abs/2406.10774}, 
}

@misc{sun2025shadowkvkvcacheshadows,
      title={ShadowKV: KV Cache in Shadows for High-Throughput Long-Context LLM Inference}, 
      author={Hanshi Sun and Li-Wen Chang and Wenlei Bao and Size Zheng and Ningxin Zheng and Xin Liu and Harry Dong and Yuejie Chi and Beidi Chen},
      year={2025},
      eprint={2410.21465},
      archivePrefix={arXiv},
      primaryClass={cs.LG},
      url={https://arxiv.org/abs/2410.21465}, 
}

@misc{zhu2025sentencekvefficientllminference,
      title={SentenceKV: Efficient LLM Inference via Sentence-Level Semantic KV Caching}, 
      author={Yuxuan Zhu and Ali Falahati and David H. Yang and Mohammad Mohammadi Amiri},
      year={2025},
      eprint={2504.00970},
      archivePrefix={arXiv},
      primaryClass={cs.CL},
      url={https://arxiv.org/abs/2504.00970}, 
}

@misc{xiao2024efficientstreaminglanguagemodels,
      title={Efficient Streaming Language Models with Attention Sinks}, 
      author={Guangxuan Xiao and Yuandong Tian and Beidi Chen and Song Han and Mike Lewis},
      year={2024},
      eprint={2309.17453},
      archivePrefix={arXiv},
      primaryClass={cs.CL},
      url={https://arxiv.org/abs/2309.17453}, 
}

@misc{zhang2023h2oheavyhitteroracleefficient,
      title={H$_2$O: Heavy-Hitter Oracle for Efficient Generative Inference of Large Language Models}, 
      author={Zhenyu Zhang and Ying Sheng and Tianyi Zhou and Tianlong Chen and Lianmin Zheng and Ruisi Cai and Zhao Song and Yuandong Tian and Christopher Ré and Clark Barrett and Zhangyang Wang and Beidi Chen},
      year={2023},
      eprint={2306.14048},
      archivePrefix={arXiv},
      primaryClass={cs.LG},
      url={https://arxiv.org/abs/2306.14048}, 
}

@misc{chen2025sepllmacceleratelargelanguage,
      title={SepLLM: Accelerate Large Language Models by Compressing One Segment into One Separator}, 
      author={Guoxuan Chen and Han Shi and Jiawei Li and Yihang Gao and Xiaozhe Ren and Yimeng Chen and Xin Jiang and Zhenguo Li and Weiyang Liu and Chao Huang},
      year={2025},
      eprint={2412.12094},
      archivePrefix={arXiv},
      primaryClass={cs.CL},
      url={https://arxiv.org/abs/2412.12094}, 
}

@book{bartlett1932remembering,
  title={Remembering: A Study in Experimental and Social Psychology},
  author={Bartlett, Frederic Charles},
  year={1932},
  publisher={Cambridge University Press},
  address={Cambridge}
}

@misc{kimiteam2025kimik15scalingreinforcement,
      title={Kimi k1.5: Scaling Reinforcement Learning with LLMs}, 
      author={Kimi Team and Angang Du and Bofei Gao and Bowei Xing and Changjiu Jiang and Cheng Chen and Cheng Li and Chenjun Xiao and Chenzhuang Du and Chonghua Liao and Chuning Tang and Congcong Wang and Dehao Zhang and Enming Yuan and Enzhe Lu and Fengxiang Tang and Flood Sung and Guangda Wei and Guokun Lai and Haiqing Guo and Han Zhu and Hao Ding and Hao Hu and Hao Yang and Hao Zhang and Haotian Yao and Haotian Zhao and Haoyu Lu and Haoze Li and Haozhen Yu and Hongcheng Gao and Huabin Zheng and Huan Yuan and Jia Chen and Jianhang Guo and Jianlin Su and Jianzhou Wang and Jie Zhao and Jin Zhang and Jingyuan Liu and Junjie Yan and Junyan Wu and Lidong Shi and Ling Ye and Longhui Yu and Mengnan Dong and Neo Zhang and Ningchen Ma and Qiwei Pan and Qucheng Gong and Shaowei Liu and Shengling Ma and Shupeng Wei and Sihan Cao and Siying Huang and Tao Jiang and Weihao Gao and Weimin Xiong and Weiran He and Weixiao Huang and Weixin Xu and Wenhao Wu and Wenyang He and Xianghui Wei and Xianqing Jia and Xingzhe Wu and Xinran Xu and Xinxing Zu and Xinyu Zhou and Xuehai Pan and Y. Charles and Yang Li and Yangyang Hu and Yangyang Liu and Yanru Chen and Yejie Wang and Yibo Liu and Yidao Qin and Yifeng Liu and Ying Yang and Yiping Bao and Yulun Du and Yuxin Wu and Yuzhi Wang and Zaida Zhou and Zhaoji Wang and Zhaowei Li and Zhen Zhu and Zheng Zhang and Zhexu Wang and Zhilin Yang and Zhiqi Huang and Zihao Huang and Ziyao Xu and Zonghan Yang and Zongyu Lin},
      year={2025},
      eprint={2501.12599},
      archivePrefix={arXiv},
      primaryClass={cs.AI},
      url={https://arxiv.org/abs/2501.12599}, 
}

@misc{aytes2025sketchofthoughtefficientllmreasoning,
      title={Sketch-of-Thought: Efficient LLM Reasoning with Adaptive Cognitive-Inspired Sketching}, 
      author={Simon A. Aytes and Jinheon Baek and Sung Ju Hwang},
      year={2025},
      eprint={2503.05179},
      archivePrefix={arXiv},
      primaryClass={cs.CL},
      url={https://arxiv.org/abs/2503.05179}, 
}

@misc{han2025tokenbudgetawarellmreasoning,
      title={Token-Budget-Aware LLM Reasoning}, 
      author={Tingxu Han and Zhenting Wang and Chunrong Fang and Shiyu Zhao and Shiqing Ma and Zhenyu Chen},
      year={2025},
      eprint={2412.18547},
      archivePrefix={arXiv},
      primaryClass={cs.CL},
      url={https://arxiv.org/abs/2412.18547}, 
}

@misc{zhang2025lightthinkerthinkingstepbystepcompression,
      title={LightThinker: Thinking Step-by-Step Compression}, 
      author={Jintian Zhang and Yuqi Zhu and Mengshu Sun and Yujie Luo and Shuofei Qiao and Lun Du and Da Zheng and Huajun Chen and Ningyu Zhang},
      year={2025},
      eprint={2502.15589},
      archivePrefix={arXiv},
      primaryClass={cs.CL},
      url={https://arxiv.org/abs/2502.15589}, 
}

@misc{yan2025inftythinkbreakinglengthlimits,
      title={InftyThink: Breaking the Length Limits of Long-Context Reasoning in Large Language Models}, 
      author={Yuchen Yan and Yongliang Shen and Yang Liu and Jin Jiang and Mengdi Zhang and Jian Shao and Yueting Zhuang},
      year={2025},
      eprint={2503.06692},
      archivePrefix={arXiv},
      primaryClass={cs.CL},
      url={https://arxiv.org/abs/2503.06692}, 
}

@misc{grattafiori2024llama3herdmodels,
      title={The Llama 3 Herd of Models}, 
      author={Aaron Grattafiori and Abhimanyu Dubey and Abhinav Jauhri and Abhinav Pandey and Abhishek Kadian and Ahmad Al-Dahle and Aiesha Letman and Akhil Mathur and Alan Schelten and Alex Vaughan and Amy Yang and Angela Fan and Anirudh Goyal and Anthony Hartshorn and Aobo Yang and Archi Mitra and Archie Sravankumar and Artem Korenev and Arthur Hinsvark and Arun Rao and Aston Zhang and Aurelien Rodriguez and Austen Gregerson and Ava Spataru and Baptiste Roziere and Bethany Biron and Binh Tang and Bobbie Chern and Charlotte Caucheteux and Chaya Nayak and Chloe Bi and Chris Marra and Chris McConnell and Christian Keller and Christophe Touret and Chunyang Wu and Corinne Wong and Cristian Canton Ferrer and Cyrus Nikolaidis and Damien Allonsius and Daniel Song and Danielle Pintz and Danny Livshits and Danny Wyatt and David Esiobu and Dhruv Choudhary and Dhruv Mahajan and Diego Garcia-Olano and Diego Perino and Dieuwke Hupkes and Egor Lakomkin and Ehab AlBadawy and Elina Lobanova and Emily Dinan and Eric Michael Smith and Filip Radenovic and Francisco Guzmán and Frank Zhang and Gabriel Synnaeve and Gabrielle Lee and Georgia Lewis Anderson and Govind Thattai and Graeme Nail and Gregoire Mialon and Guan Pang and Guillem Cucurell and Hailey Nguyen and Hannah Korevaar and Hu Xu and Hugo Touvron and Iliyan Zarov and Imanol Arrieta Ibarra and Isabel Kloumann and Ishan Misra and Ivan Evtimov and Jack Zhang and Jade Copet and Jaewon Lee and Jan Geffert and Jana Vranes and Jason Park and Jay Mahadeokar and Jeet Shah and Jelmer van der Linde and Jennifer Billock and Jenny Hong and Jenya Lee and Jeremy Fu and Jianfeng Chi and Jianyu Huang and Jiawen Liu and Jie Wang and Jiecao Yu and Joanna Bitton and Joe Spisak and Jongsoo Park and Joseph Rocca and Joshua Johnstun and Joshua Saxe and Junteng Jia and Kalyan Vasuden Alwala and Karthik Prasad and Kartikeya Upasani and Kate Plawiak and Ke Li and Kenneth Heafield and Kevin Stone and Khalid El-Arini and Krithika Iyer and Kshitiz Malik and Kuenley Chiu and Kunal Bhalla and Kushal Lakhotia and Lauren Rantala-Yeary and Laurens van der Maaten and Lawrence Chen and Liang Tan and Liz Jenkins and Louis Martin and Lovish Madaan and Lubo Malo and Lukas Blecher and Lukas Landzaat and Luke de Oliveira and Madeline Muzzi and Mahesh Pasupuleti and Mannat Singh and Manohar Paluri and Marcin Kardas and Maria Tsimpoukelli and Mathew Oldham and Mathieu Rita and Maya Pavlova and Melanie Kambadur and Mike Lewis and Min Si and Mitesh Kumar Singh and Mona Hassan and Naman Goyal and Narjes Torabi and Nikolay Bashlykov and Nikolay Bogoychev and Niladri Chatterji and Ning Zhang and Olivier Duchenne and Onur Çelebi and Patrick Alrassy and Pengchuan Zhang and Pengwei Li and Petar Vasic and Peter Weng and Prajjwal Bhargava and Pratik Dubal and Praveen Krishnan and Punit Singh Koura and Puxin Xu and Qing He and Qingxiao Dong and Ragavan Srinivasan and Raj Ganapathy and Ramon Calderer and Ricardo Silveira Cabral and Robert Stojnic and Roberta Raileanu and Rohan Maheswari and Rohit Girdhar and Rohit Patel and Romain Sauvestre and Ronnie Polidoro and Roshan Sumbaly and Ross Taylor and Ruan Silva and Rui Hou and Rui Wang and Saghar Hosseini and Sahana Chennabasappa and Sanjay Singh and Sean Bell and Seohyun Sonia Kim and Sergey Edunov and Shaoliang Nie and Sharan Narang and Sharath Raparthy and Sheng Shen and Shengye Wan and Shruti Bhosale and Shun Zhang and Simon Vandenhende and Soumya Batra and Spencer Whitman and Sten Sootla and Stephane Collot and Suchin Gururangan and Sydney Borodinsky and Tamar Herman and Tara Fowler and Tarek Sheasha and Thomas Georgiou and Thomas Scialom and Tobias Speckbacher and Todor Mihaylov and Tong Xiao and Ujjwal Karn and Vedanuj Goswami and Vibhor Gupta and Vignesh Ramanathan and Viktor Kerkez and Vincent Gonguet and Virginie Do and Vish Vogeti and Vítor Albiero and Vladan Petrovic and Weiwei Chu and Wenhan Xiong and Wenyin Fu and Whitney Meers and Xavier Martinet and Xiaodong Wang and Xiaofang Wang and Xiaoqing Ellen Tan and Xide Xia and Xinfeng Xie and Xuchao Jia and Xuewei Wang and Yaelle Goldschlag and Yashesh Gaur and Yasmine Babaei and Yi Wen and Yiwen Song and Yuchen Zhang and Yue Li and Yuning Mao and Zacharie Delpierre Coudert and Zheng Yan and Zhengxing Chen and Zoe Papakipos and Aaditya Singh and Aayushi Srivastava and Abha Jain and Adam Kelsey and Adam Shajnfeld and Adithya Gangidi and Adolfo Victoria and Ahuva Goldstand and Ajay Menon and Ajay Sharma and Alex Boesenberg and Alexei Baevski and Allie Feinstein and Amanda Kallet and Amit Sangani and Amos Teo and Anam Yunus and Andrei Lupu and Andres Alvarado and Andrew Caples and Andrew Gu and Andrew Ho and Andrew Poulton and Andrew Ryan and Ankit Ramchandani and Annie Dong and Annie Franco and Anuj Goyal and Aparajita Saraf and Arkabandhu Chowdhury and Ashley Gabriel and Ashwin Bharambe and Assaf Eisenman and Azadeh Yazdan and Beau James and Ben Maurer and Benjamin Leonhardi and Bernie Huang and Beth Loyd and Beto De Paola and Bhargavi Paranjape and Bing Liu and Bo Wu and Boyu Ni and Braden Hancock and Bram Wasti and Brandon Spence and Brani Stojkovic and Brian Gamido and Britt Montalvo and Carl Parker and Carly Burton and Catalina Mejia and Ce Liu and Changhan Wang and Changkyu Kim and Chao Zhou and Chester Hu and Ching-Hsiang Chu and Chris Cai and Chris Tindal and Christoph Feichtenhofer and Cynthia Gao and Damon Civin and Dana Beaty and Daniel Kreymer and Daniel Li and David Adkins and David Xu and Davide Testuggine and Delia David and Devi Parikh and Diana Liskovich and Didem Foss and Dingkang Wang and Duc Le and Dustin Holland and Edward Dowling and Eissa Jamil and Elaine Montgomery and Eleonora Presani and Emily Hahn and Emily Wood and Eric-Tuan Le and Erik Brinkman and Esteban Arcaute and Evan Dunbar and Evan Smothers and Fei Sun and Felix Kreuk and Feng Tian and Filippos Kokkinos and Firat Ozgenel and Francesco Caggioni and Frank Kanayet and Frank Seide and Gabriela Medina Florez and Gabriella Schwarz and Gada Badeer and Georgia Swee and Gil Halpern and Grant Herman and Grigory Sizov and Guangyi and Zhang and Guna Lakshminarayanan and Hakan Inan and Hamid Shojanazeri and Han Zou and Hannah Wang and Hanwen Zha and Haroun Habeeb and Harrison Rudolph and Helen Suk and Henry Aspegren and Hunter Goldman and Hongyuan Zhan and Ibrahim Damlaj and Igor Molybog and Igor Tufanov and Ilias Leontiadis and Irina-Elena Veliche and Itai Gat and Jake Weissman and James Geboski and James Kohli and Janice Lam and Japhet Asher and Jean-Baptiste Gaya and Jeff Marcus and Jeff Tang and Jennifer Chan and Jenny Zhen and Jeremy Reizenstein and Jeremy Teboul and Jessica Zhong and Jian Jin and Jingyi Yang and Joe Cummings and Jon Carvill and Jon Shepard and Jonathan McPhie and Jonathan Torres and Josh Ginsburg and Junjie Wang and Kai Wu and Kam Hou U and Karan Saxena and Kartikay Khandelwal and Katayoun Zand and Kathy Matosich and Kaushik Veeraraghavan and Kelly Michelena and Keqian Li and Kiran Jagadeesh and Kun Huang and Kunal Chawla and Kyle Huang and Lailin Chen and Lakshya Garg and Lavender A and Leandro Silva and Lee Bell and Lei Zhang and Liangpeng Guo and Licheng Yu and Liron Moshkovich and Luca Wehrstedt and Madian Khabsa and Manav Avalani and Manish Bhatt and Martynas Mankus and Matan Hasson and Matthew Lennie and Matthias Reso and Maxim Groshev and Maxim Naumov and Maya Lathi and Meghan Keneally and Miao Liu and Michael L. Seltzer and Michal Valko and Michelle Restrepo and Mihir Patel and Mik Vyatskov and Mikayel Samvelyan and Mike Clark and Mike Macey and Mike Wang and Miquel Jubert Hermoso and Mo Metanat and Mohammad Rastegari and Munish Bansal and Nandhini Santhanam and Natascha Parks and Natasha White and Navyata Bawa and Nayan Singhal and Nick Egebo and Nicolas Usunier and Nikhil Mehta and Nikolay Pavlovich Laptev and Ning Dong and Norman Cheng and Oleg Chernoguz and Olivia Hart and Omkar Salpekar and Ozlem Kalinli and Parkin Kent and Parth Parekh and Paul Saab and Pavan Balaji and Pedro Rittner and Philip Bontrager and Pierre Roux and Piotr Dollar and Polina Zvyagina and Prashant Ratanchandani and Pritish Yuvraj and Qian Liang and Rachad Alao and Rachel Rodriguez and Rafi Ayub and Raghotham Murthy and Raghu Nayani and Rahul Mitra and Rangaprabhu Parthasarathy and Raymond Li and Rebekkah Hogan and Robin Battey and Rocky Wang and Russ Howes and Ruty Rinott and Sachin Mehta and Sachin Siby and Sai Jayesh Bondu and Samyak Datta and Sara Chugh and Sara Hunt and Sargun Dhillon and Sasha Sidorov and Satadru Pan and Saurabh Mahajan and Saurabh Verma and Seiji Yamamoto and Sharadh Ramaswamy and Shaun Lindsay and Shaun Lindsay and Sheng Feng and Shenghao Lin and Shengxin Cindy Zha and Shishir Patil and Shiva Shankar and Shuqiang Zhang and Shuqiang Zhang and Sinong Wang and Sneha Agarwal and Soji Sajuyigbe and Soumith Chintala and Stephanie Max and Stephen Chen and Steve Kehoe and Steve Satterfield and Sudarshan Govindaprasad and Sumit Gupta and Summer Deng and Sungmin Cho and Sunny Virk and Suraj Subramanian and Sy Choudhury and Sydney Goldman and Tal Remez and Tamar Glaser and Tamara Best and Thilo Koehler and Thomas Robinson and Tianhe Li and Tianjun Zhang and Tim Matthews and Timothy Chou and Tzook Shaked and Varun Vontimitta and Victoria Ajayi and Victoria Montanez and Vijai Mohan and Vinay Satish Kumar and Vishal Mangla and Vlad Ionescu and Vlad Poenaru and Vlad Tiberiu Mihailescu and Vladimir Ivanov and Wei Li and Wenchen Wang and Wenwen Jiang and Wes Bouaziz and Will Constable and Xiaocheng Tang and Xiaojian Wu and Xiaolan Wang and Xilun Wu and Xinbo Gao and Yaniv Kleinman and Yanjun Chen and Ye Hu and Ye Jia and Ye Qi and Yenda Li and Yilin Zhang and Ying Zhang and Yossi Adi and Youngjin Nam and Yu and Wang and Yu Zhao and Yuchen Hao and Yundi Qian and Yunlu Li and Yuzi He and Zach Rait and Zachary DeVito and Zef Rosnbrick and Zhaoduo Wen and Zhenyu Yang and Zhiwei Zhao and Zhiyu Ma},
      year={2024},
      eprint={2407.21783},
      archivePrefix={arXiv},
      primaryClass={cs.AI},
      url={https://arxiv.org/abs/2407.21783}, 
}

@misc{qwen2025qwen25technicalreport,
      title={Qwen2.5 Technical Report}, 
      author={Qwen and : and An Yang and Baosong Yang and Beichen Zhang and Binyuan Hui and Bo Zheng and Bowen Yu and Chengyuan Li and Dayiheng Liu and Fei Huang and Haoran Wei and Huan Lin and Jian Yang and Jianhong Tu and Jianwei Zhang and Jianxin Yang and Jiaxi Yang and Jingren Zhou and Junyang Lin and Kai Dang and Keming Lu and Keqin Bao and Kexin Yang and Le Yu and Mei Li and Mingfeng Xue and Pei Zhang and Qin Zhu and Rui Men and Runji Lin and Tianhao Li and Tianyi Tang and Tingyu Xia and Xingzhang Ren and Xuancheng Ren and Yang Fan and Yang Su and Yichang Zhang and Yu Wan and Yuqiong Liu and Zeyu Cui and Zhenru Zhang and Zihan Qiu},
      year={2025},
      eprint={2412.15115},
      archivePrefix={arXiv},
      primaryClass={cs.CL},
      url={https://arxiv.org/abs/2412.15115}, 
}

@misc{bespoke_stratos,  
    author = {Bespoke Labs},  
    title = {Bespoke-Stratos: The unreasonable effectiveness of reasoning distillation},  
    howpublished = {https://www.bespokelabs.ai/blog/bespoke-stratos-the-unreasonable-effectiveness-of-reasoning-distillation},  
    note = {Accessed: 2025-01-22},  
    year = {2025}
}

@misc{hu2021loralowrankadaptationlarge,
      title={LoRA: Low-Rank Adaptation of Large Language Models}, 
      author={Edward J. Hu and Yelong Shen and Phillip Wallis and Zeyuan Allen-Zhu and Yuanzhi Li and Shean Wang and Lu Wang and Weizhu Chen},
      year={2021},
      eprint={2106.09685},
      archivePrefix={arXiv},
      primaryClass={cs.CL},
      url={https://arxiv.org/abs/2106.09685}, 
}

@misc{openai2024gpt4technicalreport,
      title={GPT-4 Technical Report}, 
      author={OpenAI and Josh Achiam and Steven Adler and Sandhini Agarwal and Lama Ahmad and Ilge Akkaya and Florencia Leoni Aleman and Diogo Almeida and Janko Altenschmidt and Sam Altman and Shyamal Anadkat and Red Avila and Igor Babuschkin and Suchir Balaji and Valerie Balcom and Paul Baltescu and Haiming Bao and Mohammad Bavarian and Jeff Belgum and Irwan Bello and Jake Berdine and Gabriel Bernadett-Shapiro and Christopher Berner and Lenny Bogdonoff and Oleg Boiko and Madelaine Boyd and Anna-Luisa Brakman and Greg Brockman and Tim Brooks and Miles Brundage and Kevin Button and Trevor Cai and Rosie Campbell and Andrew Cann and Brittany Carey and Chelsea Carlson and Rory Carmichael and Brooke Chan and Che Chang and Fotis Chantzis and Derek Chen and Sully Chen and Ruby Chen and Jason Chen and Mark Chen and Ben Chess and Chester Cho and Casey Chu and Hyung Won Chung and Dave Cummings and Jeremiah Currier and Yunxing Dai and Cory Decareaux and Thomas Degry and Noah Deutsch and Damien Deville and Arka Dhar and David Dohan and Steve Dowling and Sheila Dunning and Adrien Ecoffet and Atty Eleti and Tyna Eloundou and David Farhi and Liam Fedus and Niko Felix and Simón Posada Fishman and Juston Forte and Isabella Fulford and Leo Gao and Elie Georges and Christian Gibson and Vik Goel and Tarun Gogineni and Gabriel Goh and Rapha Gontijo-Lopes and Jonathan Gordon and Morgan Grafstein and Scott Gray and Ryan Greene and Joshua Gross and Shixiang Shane Gu and Yufei Guo and Chris Hallacy and Jesse Han and Jeff Harris and Yuchen He and Mike Heaton and Johannes Heidecke and Chris Hesse and Alan Hickey and Wade Hickey and Peter Hoeschele and Brandon Houghton and Kenny Hsu and Shengli Hu and Xin Hu and Joost Huizinga and Shantanu Jain and Shawn Jain and Joanne Jang and Angela Jiang and Roger Jiang and Haozhun Jin and Denny Jin and Shino Jomoto and Billie Jonn and Heewoo Jun and Tomer Kaftan and Łukasz Kaiser and Ali Kamali and Ingmar Kanitscheider and Nitish Shirish Keskar and Tabarak Khan and Logan Kilpatrick and Jong Wook Kim and Christina Kim and Yongjik Kim and Jan Hendrik Kirchner and Jamie Kiros and Matt Knight and Daniel Kokotajlo and Łukasz Kondraciuk and Andrew Kondrich and Aris Konstantinidis and Kyle Kosic and Gretchen Krueger and Vishal Kuo and Michael Lampe and Ikai Lan and Teddy Lee and Jan Leike and Jade Leung and Daniel Levy and Chak Ming Li and Rachel Lim and Molly Lin and Stephanie Lin and Mateusz Litwin and Theresa Lopez and Ryan Lowe and Patricia Lue and Anna Makanju and Kim Malfacini and Sam Manning and Todor Markov and Yaniv Markovski and Bianca Martin and Katie Mayer and Andrew Mayne and Bob McGrew and Scott Mayer McKinney and Christine McLeavey and Paul McMillan and Jake McNeil and David Medina and Aalok Mehta and Jacob Menick and Luke Metz and Andrey Mishchenko and Pamela Mishkin and Vinnie Monaco and Evan Morikawa and Daniel Mossing and Tong Mu and Mira Murati and Oleg Murk and David Mély and Ashvin Nair and Reiichiro Nakano and Rajeev Nayak and Arvind Neelakantan and Richard Ngo and Hyeonwoo Noh and Long Ouyang and Cullen O'Keefe and Jakub Pachocki and Alex Paino and Joe Palermo and Ashley Pantuliano and Giambattista Parascandolo and Joel Parish and Emy Parparita and Alex Passos and Mikhail Pavlov and Andrew Peng and Adam Perelman and Filipe de Avila Belbute Peres and Michael Petrov and Henrique Ponde de Oliveira Pinto and Michael and Pokorny and Michelle Pokrass and Vitchyr H. Pong and Tolly Powell and Alethea Power and Boris Power and Elizabeth Proehl and Raul Puri and Alec Radford and Jack Rae and Aditya Ramesh and Cameron Raymond and Francis Real and Kendra Rimbach and Carl Ross and Bob Rotsted and Henri Roussez and Nick Ryder and Mario Saltarelli and Ted Sanders and Shibani Santurkar and Girish Sastry and Heather Schmidt and David Schnurr and John Schulman and Daniel Selsam and Kyla Sheppard and Toki Sherbakov and Jessica Shieh and Sarah Shoker and Pranav Shyam and Szymon Sidor and Eric Sigler and Maddie Simens and Jordan Sitkin and Katarina Slama and Ian Sohl and Benjamin Sokolowsky and Yang Song and Natalie Staudacher and Felipe Petroski Such and Natalie Summers and Ilya Sutskever and Jie Tang and Nikolas Tezak and Madeleine B. Thompson and Phil Tillet and Amin Tootoonchian and Elizabeth Tseng and Preston Tuggle and Nick Turley and Jerry Tworek and Juan Felipe Cerón Uribe and Andrea Vallone and Arun Vijayvergiya and Chelsea Voss and Carroll Wainwright and Justin Jay Wang and Alvin Wang and Ben Wang and Jonathan Ward and Jason Wei and CJ Weinmann and Akila Welihinda and Peter Welinder and Jiayi Weng and Lilian Weng and Matt Wiethoff and Dave Willner and Clemens Winter and Samuel Wolrich and Hannah Wong and Lauren Workman and Sherwin Wu and Jeff Wu and Michael Wu and Kai Xiao and Tao Xu and Sarah Yoo and Kevin Yu and Qiming Yuan and Wojciech Zaremba and Rowan Zellers and Chong Zhang and Marvin Zhang and Shengjia Zhao and Tianhao Zheng and Juntang Zhuang and William Zhuk and Barret Zoph},
      year={2024},
      eprint={2303.08774},
      archivePrefix={arXiv},
      primaryClass={cs.CL},
      url={https://arxiv.org/abs/2303.08774}, 
}

@article{li2024snapkv,
  title={Snapkv: Llm knows what you are looking for before generation},
  author={Li, Yuhong and Huang, Yingbing and Yang, Bowen and Venkitesh, Bharat and Locatelli, Acyr and Ye, Hanchen and Cai, Tianle and Lewis, Patrick and Chen, Deming},
  journal={Advances in Neural Information Processing Systems},
  volume={37},
  pages={22947--22970},
  year={2024}
}

@article{saxena2024eigen,
  title={Eigen attention: Attention in low-rank space for kv cache compression},
  author={Saxena, Utkarsh and Saha, Gobinda and Choudhary, Sakshi and Roy, Kaushik},
  journal={arXiv preprint arXiv:2408.05646},
  year={2024}
}

@misc{zhu2025ojakvcontextawareonlinelowrank,
      title={OjaKV: Context-Aware Online Low-Rank KV Cache Compression with Oja's Rule}, 
      author={Yuxuan Zhu and David H. Yang and Mohammad Mohammadi Amiri and Keerthiram Murugesan and Tejaswini Pedapati and Pin-Yu Chen},
      year={2025},
      eprint={2509.21623},
      archivePrefix={arXiv},
      primaryClass={cs.CL},
      url={https://arxiv.org/abs/2509.21623}, 
}

@article{shen2025codi,
  title={Codi: Compressing chain-of-thought into continuous space via self-distillation},
  author={Shen, Zhenyi and Yan, Hanqi and Zhang, Linhai and Hu, Zhanghao and Du, Yali and He, Yulan},
  journal={arXiv preprint arXiv:2502.21074},
  year={2025}
}

@article{hao2024training,
  title={Training large language models to reason in a continuous latent space},
  author={Hao, Shibo and Sukhbaatar, Sainbayar and Su, DiJia and Li, Xian and Hu, Zhiting and Weston, Jason and Tian, Yuandong},
  journal={arXiv preprint arXiv:2412.06769},
  year={2024}
}

@article{https://doi.org/10.13140/rg.2.2.28167.37282,
  doi = {10.13140/RG.2.2.28167.37282},
  
  url = {https://rgdoi.net/10.13140/RG.2.2.28167.37282},
  
  author = {{Zirui Liu} and {Jiayi Yuan} and {Hongye Jin} and {Shaochen Zhong} and {Zhaozhuo Xu} and Braverman, Vladimir and {Beidi Chen} and Hu, Xia},
  
  language = {en},
  
  title = {KIVI : Plug-and-play 2bit KV Cache Quantization with Streaming Asymmetric Quantization},
  
  publisher = {Unpublished},
  
  year = {2023}
}

@misc{hooper2025kvquant10millioncontext,
      title={KVQuant: Towards 10 Million Context Length LLM Inference with KV Cache Quantization}, 
      author={Coleman Hooper and Sehoon Kim and Hiva Mohammadzadeh and Michael W. Mahoney and Yakun Sophia Shao and Kurt Keutzer and Amir Gholami},
      year={2025},
      eprint={2401.18079},
      archivePrefix={arXiv},
      primaryClass={cs.LG},
      url={https://arxiv.org/abs/2401.18079}, 
}

\appendix

\newpage

\section{Appendix}

\subsection{Summarization}
\label{appendix:summarization}
ZoomR uses a Llama3-70B model to generate summaries and augment the bespoke-17K dataset.\ Specifically, we use the summarization prompt shown in Figure \ref{appendix_figure_summary}.\ Since summaries can be generated independently, this process can be greatly accelerated with the use of larger batch sizes and API calls.\ Once we collect all the summaries, these are then inserted back into the original dataset between special tokens <|begin\_of\_summary|> and <|end\_of\_summary|>, as illustrated in Figure \ref{appendix_figure_summary1}. Fine-tuning hyperparameters are shown in Table \ref{lora_table}.

\begin{table}[h]
\centering
\caption{LoRA Fine-tuning Hyperparameters}
\label{tab:lora_hyperparameters}
\begin{tabular}{ll}
\hline
\textbf{Parameter} & \textbf{Value} \\
\hline
LoRA Rank ($r$) & 16 \\
LoRA Alpha ($\alpha$) & 32 \\
Number of Epochs & 3 \\
Learning Rate & $1 \times 10^{-4}$ \\
Batch Size & 1 \\
Gradient Accumulation Steps & 4 \\
Maximum Sequence Length & 16,384 \\
\hline
\end{tabular}
\label{lora_table}
\end{table}

\subsection{Algorithm Details}
\label{appendix_alg}
Algorithm \ref{alg:ZoomR} shows the full algorithm for ZoomR.

\subsection{Semantic Boundaries}
\label{app:sem_bound}
In our methodology, the semantic boundaries are determined by a preset list of punctuation tokens.\ The specific tokens used are \verb|.|, \verb|!|, \verb|?|, \verb|...|, \verb|:|, \verb|\n\n|, and \verb|\n\n\n|.\ The average recompute interval occurs after approximately 35 tokens, which matches the expected sentence length.

\subsection{ZoomR Generated Example}
ZoomR generates summaries dynamically during reasoning. Figure \ref{appendix_figure_summary2} shows part of the reasoning thoughts from a challenging AIME2025 problem.

\begin{figure*}[htbp]
\begin{tcolorbox}[
    colback=black!5,      
    colframe=black!75,    
    title=Summarization Prompt,
    fonttitle=\bfseries,  
    arc=2mm,              
    boxrule=1pt           
]
\begin{verbatim}
System Prompt: You are a reasoning trace compressor. Your goal is to 
compress intermediate reasoning steps into minimal but faithful summaries.

Instruction:
"You are compressing a single paragraph from a chain-of-thought reasoning"
"trace into a short summary token sequence.\ Use the prior context to"
"understand what the current step is doing, but summarize only the"
"current paragraph.\ Context: " + {context} + "\n"
"Compress the following paragraph into a single short phrase (10-12 tokens)"
"that captures the key logical conclusion or operation performed."
"Prioritize preserving: mathematical results, logical deductions, and"
"intermediate conclusions.\ You may omit redundant reasoning steps."
"Respond in JSON with the compressed summary in 'results'.\n"
"Paragraph: \n" + {paragraph}
\end{verbatim}
\end{tcolorbox}
\caption{Prompt used to generate summaries}
\label{appendix_figure_summary}
\end{figure*}

\begin{figure*}[htbp]
\begin{tcolorbox}[
    colback=black!5,      
    colframe=black!75,    
    title=Augmented Dataset Example,
    fonttitle=\bfseries,  
    arc=2mm,              
    boxrule=1pt           
]
\textbf{Question:} Return your final response within \verb|\boxed{}|.\ A quadrilateral has vertices at (0,1), (3,4), (4,3) and (3,0).\ Its perimeter can be expressed in the form $a\sqrt{2}+b\sqrt{10}$ with $a$ and $b$ integers.\ What is the sum of $a$ and $b$?

\vspace{1em}
\textbf{Thoughts:} <|begin\_of\_thought|> Okay, so I have this problem where I need to find the perimeter of a quadrilateral with vertices at (0,1), (3,4), (4,3), and (3,0).\ The perimeter is supposed to be expressed in the form $a\sqrt{2} + b\sqrt{10}$, where a and b are integers, and then I have to find the sum of a and b.\ Alright, let me start by understanding how to approach this.

First, the perimeter is the sum of the lengths of all four sides.\ I need to find the distance between each pair of consecutive vertices.\ The distance formula between two points $(x_1, y_1)$ and $(x_2, y_2)$ is $\sqrt{(x_2 - x_1)^2 + (y_2 - y_1)^2}$.

Let the vertices be A(0,1), B(3,4), C(4,3), D(3,0).\ The sides are AB, BC, CD, and DA.

<|begin\_of\_summary|>$\text{Perimeter} = AB + BC + CD + DA$<|end\_of\_summary|>

I'll calculate each side one by one.

Starting with AB: from (0,1) to (3,4).\ The length AB is $\sqrt{(3-0)^2 + (4-1)^2}$ = $\sqrt{3^2 + 3^2} = \sqrt{9 + 9} = \sqrt{18}$.

This can be simplified: $\sqrt{18} = \sqrt{9 \times 2} = 3\sqrt{2}$.\ So AB is $3\sqrt{2}$.

Next, BC: from (3,4) to (4,3).\ The length BC is $\sqrt{(4-3)^2 + (3-4)^2}$ = $\sqrt{1^2 + (-1)^2} = \sqrt{1 + 1} = \sqrt{2}$.

<|begin\_of\_summary|>$AB = 3\sqrt{2}$, $BC = \sqrt{2}$<|end\_of\_summary|>

Third side is CD: from (4,3) to (3,0).\ The length CD is $\sqrt{(3-4)^2 + (0-3)^2}$ = $\sqrt{(-1)^2 + (-3)^2} = \sqrt{1 + 9} = \sqrt{10}$.

Last side is DA: from (3,0) back to (0,1).\ The length DA is $\sqrt{(0-3)^2 + (1-0)^2}$ = $\sqrt{(-3)^2 + 1^2} = \sqrt{9 + 1} = \sqrt{10}$.

Now, let's add up all these lengths: AB + BC + CD + DA = $3\sqrt{2} + \sqrt{2} + \sqrt{10} + \sqrt{10}$.

<|begin\_of\_summary|>$AB = 3\sqrt{2}$, $BC = \sqrt{2}$, $CD = \sqrt{10}$, $DA = \sqrt{10}$<|end\_of\_summary|>

Combining like terms: $(3\sqrt{2} + \sqrt{2}) + (\sqrt{10} + \sqrt{10}) = 4\sqrt{2} + 2\sqrt{10}$.

So the total perimeter is $4\sqrt{2} + 2\sqrt{10}$.

$\cdots$
\end{tcolorbox}

\caption{An example of a segment of reasoning process with intermediate summaries.}
\label{appendix_figure_summary1}
\end{figure*}

\begin{algorithm*}[htbp]
\caption{ZoomR}
\label{alg:ZoomR}
\begin{algorithmic}[1]
\State \textbf{Require:} Prompt indices $\mathcal{I}_p$, per-head top-$k$, consensus count $c$, recent window size $w$.
\State \textbf{Define:} Regular segments $\{\mathcal{R}_i\}$ and summary segments $\{\mathcal{S}_i\}$.
\Statex
\Statex \textbf{Prefill:} Initialize KV cache using prompt $P$.
\Statex \textbf{Decoding:}
\For{each step $t = 1, \dots, N_g$}
        \For{each layer $l=1, \dots, N_L$} \Comment{Per-head importance scoring and voting}
            \For{each head $h=1, \dots, H$}
            \State Append $\boldsymbol{k}_t^{(l,h)}$ and $\boldsymbol{v}_t^{(l,h)}$ to KV cache
            \If{a new summary $\mathcal{S}_{N_t}$ is complete}
                \State $\bar{\boldsymbol{k}}_{N_t}^{(l,h)} \gets \frac{1}{|\mathcal{S}_{N_t}|} \sum_{j \in \mathcal{S}_{N_t}} \boldsymbol{k}_j^{(l,h)}$. \Comment{Compute mean summary key}
            \EndIf
                \State $\alpha_i^{(l,h)} \gets (\boldsymbol{q}_t^{(l,h)})^{\top} \bar{\boldsymbol{k}}_i^{(l,h)}$ for all $i$.
                \State $\mathcal{I}_{\text{top-k}}^{(l,h)} \gets \underset{i}{\arg\text{top-k}}(\alpha_i^{(l,h)})$.
            \EndFor
        \EndFor
        \If{at a semantic boundary}
        \State $\mathcal{I}_{\text{all}} \gets \bigcup_{l,h} \mathcal{I}_{\text{top-k}}^{(l,h)}$. \Comment{Global aggregation and consensus}
        \State For each $i \in \mathcal{I}_{\text{all}}$, compute $v_i = \sum_{l,h} \mathbb{I}(i \in \mathcal{I}_{\text{top-k}}^{(l,h)})$.
        \State $\mathcal{I}_c \gets \underset{i \in \mathcal{I}_{\text{all}}}{\arg\text{top-c}}(v_i)$.
        \State $\mathcal{I}_s \gets \mathcal{I}_{\text{all}} \setminus \mathcal{I}_c$.
        
        \State $\mathcal{I}_w \gets \{t-w+1, \dots, t-1\}$.
        \State $\mathcal{I}_{f} \gets \mathcal{I}_p \cup \mathcal{I}_w \cup \left(\bigcup_{i \in \mathcal{I}_{c}} \mathcal{R}_i\right) \cup \left(\bigcup_{i \in \mathcal{I}_{s}} \mathcal{S}_i\right)$. \Comment{Construct final multi-granularity context}
    \EndIf
\EndFor
\end{algorithmic}
\end{algorithm*}

\begin{figure*}[htbp]
\begin{tcolorbox}[
    colback=black!5,      
    colframe=black!75,    
    title=ZoomR Example,
    fonttitle=\bfseries,  
    arc=2mm,              
    boxrule=1pt           
]
\textbf{Question:}
The 9 members of a baseball team went to an ice cream parlor after their game. Each player had a single-scoop cone of chocolate, vanilla, or strawberry ice cream. At least one player chose each flavor, and the number of players who chose chocolate was greater than the number of players who chose vanilla, which was greater than the number of players who chose strawberry. Let $N$ be the number of different assignments of flavors to players that meet these conditions. Find the remainder when $N$ is divided by 1000.

\vspace{1em}
\textbf{Thoughts:}
\texttt{<|begin\_of\_thought|>} Okay, so I need to solve this problem about assigning ice cream flavors to a baseball team. Let me read the problem again carefully to make sure I understand all the conditions.

There are 9 team members, each getting a single scoop of chocolate, vanilla, or strawberry ice cream. The conditions are:
\begin{enumerate}
    \item Each flavor is chosen by at least one player.
    \item The number of players who chose chocolate is greater than those who chose vanilla, which in turn is greater than those who chose strawberry. So, $Ch > Van > Str$.
\end{enumerate}

We need to find the number of different assignments (flavorings) that meet these conditions, then find $N \pmod{1000}$.

Hmm. Let me break this down. So, each player has 3 choices: $C$, $V$, or $S$. But with constraints on the counts. The key constraints are:
\begin{itemize}
    \item All three flavors are used at least once.
    \item The counts satisfy $Ch > Van > Str$.
\end{itemize}

So, variables:

Let $C$ = number of chocolate, $V$ = vanilla, $S$ = strawberry.

Given that $C > V > S$, and $C + V + S = 9$. Also, each of $C, V, S \geq 1$.
\texttt{<|begin\_of\_summary|>} $C > V > S$, $C + V + S = 9$ \texttt{<|end\_of\_summary|>}
We need to find the number of ordered triples $(C, V, S)$ that satisfy these conditions, and then multiply by the number of ways to assign these counts to the 9 players, considering the different permutations.

Wait, no. Actually, the assignments are different distributions of the flavors to the players. So, for each valid combination $(C, V, S)$, the number of ways is the multinomial coefficient: $\frac{9!}{C! V! S!}$. Then, sum this over all valid triples $(C, V, S)$ where $C > V > S \geq 1$ and $C + V + S = 9$.

But first, we need to find all possible triples $(C, V, S)$ that satisfy the inequalities and sum to 9. Let me first figure out all possible triples $(C, V, S)$ with $C > V > S \geq 1$ and $C + V + S = 9$.

Since $S \geq 1$, $V \geq 1$, $C \geq 1$, and $C > V > S$.

Let me start by considering possible values for $S$. Since $S$ must be at least 1, and $V$ must be at least $S + 1$, and $C$ must be at least $V + 1$. So, $S$ can be from 1 up to some maximum value where these inequalities can still hold.

Let me try to find the possible values of $S$.
\texttt{<|begin\_of\_summary|>} $S \geq 1, V \geq S + 1, C \geq V + 1$ \texttt{<|end\_of\_summary|>}
Let’s denote $S = k$. Then, $V$ must be at least $k + 1$, and $C$ must be at least $V + 1 = k + 2$. So, $C \geq k + 2$.

$\cdots$

Thus, the remainder when \( N \) is divided by 1000 is \(\boxed{16}\).
\end{tcolorbox}
\caption{An example from AIME2025. ZoomR generated thoughts with summarization.}
\label{appendix_figure_summary2}
\end{figure*}

\end{document}